\documentclass[lettersize,journal]{IEEEtran}
\usepackage{amsmath,amsfonts}
\usepackage{algorithmic}
\usepackage{orcidlink}
\usepackage{array}
\usepackage[caption=false,font=normalsize,labelfont=sf,textfont=sf]{subfig}
\usepackage{textcomp}
\usepackage{stfloats}
\usepackage{url}
\usepackage{verbatim}
\usepackage{graphicx,multirow,booktabs,comment}
\def\BibTeX{{\rm B\kern-.05em{\sc i\kern-.025em b}\kern-.08em
    T\kern-.1667em\lower.7ex\hbox{E}\kern-.125emX}}
\usepackage{balance}
\usepackage{amsmath,amsthm,amssymb}
\usepackage{comment,times,setspace}
\usepackage[linesnumbered,ruled]{algorithm2e}
\usepackage{bm}
\usepackage{float}

\usepackage{xcolor}
\usepackage[ruled,linesnumbered]{algorithm2e}

\SetCommentSty{mycommfont}

\usepackage{xr}
\newtheorem{theorem}{Theorem}
\newtheorem{remark}{Remark}
\newtheorem{lemma}{Lemma}

\newtheorem{prop}{Proposition}

\newtheorem{assumption}{Assumption}
\newtheorem{definition}{Definition}
\newtheorem{example}{Example}

\def\spacingset#1{\renewcommand{\baselinestretch}%
{#1}\small\normalsize} \spacingset{1}

\newcommand{\argmin}{\operatorname*{\arg\min}}
\newcommand{\argmax}{\operatorname*{\arg\max}}

\newcommand{\E}{\operatorname{E}} 
\newcommand{\Var}{\operatorname{Var}}

\newcommand {\hpz} {\tilde{p}_{\scalebox{1}{$\scriptscriptstyle \bm{Z}$}}}
\newcommand {\hF} {\tilde{F}}
\newcommand {\Fz} {F_{\scalebox{1}{$\scriptscriptstyle \bm{Z}$}}}
\newcommand {\hFz} {\tilde F_{\scalebox{1}{$\scriptscriptstyle \bm{Z}$}}}
\newcommand {\pz} {p_{\scalebox{1}{$\scriptscriptstyle \bm{Z}$}}}

\newcommand{\mb}{\boldsymbol}
\newcommand{\mc}{\mathcal}

\usepackage{stmaryrd}

\newcommand{ \brac }[1]{\left[ #1 \right]}
\newcommand{ \paren }[1]{ \left( #1 \right) }

\newcommand{\wt}{\widetilde}

\newcommand{\T}{\intercal}

\newcommand{\Y}{\bm Y}
\newcommand{\X}{\bm X}
\newcommand{\Z}{\bm Z}

\newcommand{\V}{\bm V}
\newcommand{\z}{\bm z}
\newcommand{\x}{\bm x}
\newcommand{\e}{\bm e}
\newcommand{\U}{\bm U}

\newcommand{\abs}[1]{\left| #1 \right|}

\newcommand{\deq}{\buildrel d \over =} 

\begin{document}

 \title{Novel Uncertainty Quantification through Perturbation-Assisted Sample Synthesis
 \thanks{This work was supported in part by NSF grant DMS-1952539, and NIH grants 
R01AG069895, R01AG065636, R01AG074858, U01AG073079. (Corresponding author: Xiaotong Shen.)}}

\author{
Yifei Liu, Rex Shen, Xiaotong Shen\textsuperscript{\orcidlink{0000-0003-1300-1451}}
  \thanks{Yifei Liu is with the School of Statistics, University of Minnesota, MN, 55455 USA 
  (email: liu00980@umn.edu).}
\thanks{Rex Shen is with the Department of Statistics, Stanford University, CA, 94305 
USA (email:rshen0@stanford.edu).} \thanks{Xiaotong Shen is with the School of Statistics, University of Minnesota, MN, 55455 USA (email: xshen@umn.edu).}
}

\maketitle


\begin{abstract}
This paper introduces a novel Perturbation-Assisted Inference (PAI) framework utilizing synthetic data generated by the Perturbation-Assisted Sample Synthesis (PASS) method. The framework focuses on uncertainty quantification in complex data scenarios, particularly involving unstructured data while utilizing deep learning models. On one hand,  PASS employs a generative model to create synthetic data that closely mirrors raw data while preserving its rank properties through data perturbation, thereby enhancing data diversity and bolstering privacy. By incorporating knowledge transfer from large pre-trained generative models, PASS enhances estimation accuracy, yielding refined distributional estimates of various statistics via Monte Carlo experiments. On the other hand, PAI boasts its statistically guaranteed validity. In pivotal inference, it enables precise conclusions even without prior knowledge of the pivotal's distribution. In non-pivotal situations, we enhance the reliability of synthetic data generation by training it with an independent holdout sample. We demonstrate the effectiveness of PAI in advancing uncertainty quantification in complex, data-driven tasks by applying it to diverse areas such as image synthesis, sentiment word analysis, multimodal inference, and the construction of prediction intervals.
\end{abstract}



\section{Introduction}

 Uncertainty quantification is pivotal in scientific exploration and drawing reliable conclusions from data, especially working with complex modeling techniques such as deep neural networks. Despite recent advancements showcasing the potential of Artificial Intelligence in facilitating data-driven discoveries, a reproducibility crisis has emerged in various fields, including biomedicine and social sciences, occasionally leading to false discoveries \cite{gibney2022ai}. A key issue contributing to this crisis is the lack of methods for quantifying uncertainty in over-parametrized models, like neural networks, prioritizing predictive accuracy using many non-learnable parameters such as hyperparameters. As a result, these studies may become exaggerated and irreproducible. To address these challenges, we develop a generative inference framework designed to provide uncertainty quantification for data of any type.

Diverse methodologies for uncertainty quantification are prevalent in the literature. Approaches such as those in \cite{nemani2023uncertainty, kong2023uncertainty, abdar2021review, lakshminarayanan2017simple} evaluate the predictive model's outcome uncertainty,  with broad applications spanning adversarial attacks to anomaly detection \cite{song2017pixeldefend, liang2022gmmseg}. Furthermore, studies \cite{kuhn2023semantic, lin2023generating} investigate uncertainty within large language models for question-answering tasks. Nevertheless, prevailing metrics like negative log-likelihood often forgo solid statistical foundations, such as confidence or probability assertions, within the framework of statistical inference.

In statistical inference, the quantification of uncertainty is imperative. Classical techniques like Bootstrap \cite{efron1992bootstrap} solve conventional statistical problems. 
Yet, uncertainty in complex models, particularly those involving deep networks and unstructured data, as indicated by \cite{kong2023uncertainty}, remains less explored. 
The conformal inference method \cite{lei2018distribution, angelopoulos2021gentle} offers a practical tool for valid uncertainty quantification. However, its effectiveness is significantly influenced by the underlying prediction model and the selection of the conformal score, which can lead to overly cautious inferential outcomes.
With recent progress, such as \cite{dai2024significance}'s introduction of hypothesis testing for feature significance using asymptotic methods, a comprehensive examination of statistical uncertainty quantification becomes 
imperative. Our focus herein is statistical inference, specifically hypothesis testing, which quantifies the uncertainty of a hypothesis test's outcome or conveys a confidence declaration concerning prediction uncertainty, as detailed in Section \ref{sec:deep}.

This paper introduces the novel Perturbation-Assisted Inference (PAI) framework that employs Perturbation-Assisted Sample Synthesis (PASS) as its core generator, ensuring validity as if we had conducted Monte Carlo (MC) simulations with a known data-generating distribution. To clarify the core concept of our approach, envision statistics computed via a machine learning or statistical technique on a training data set. These statistics may embody a predicted outcome in supervised learning or a test statistic in hypothesis testing. By generating multiple iterations of these statistics on synthetic data that emulate the original data's distribution, we gauge the variability of these statistics across data sets with an analogous distribution to the original by applying the same analytical method. PASS generates these synthetic data sets, while PAI procures reliable inferences from them, employing Monte Carlo techniques.

PASS synthesizes data that mirrors the original data closely, encompassing both tabular and unstructured data such as gene expressions and text. Its distinct edge is in harnessing pre-trained generative models to heighten generation precision. With an emphasis on inference, PASS augments synthetic data diversity and privacy through data perturbation, retaining the original sample's ranks, which supports personalization and data amalgamation \cite{shen2022data}. Through neural networks, PASS maps a base distribution into a  target one, drawing from the round-trip transformation strategy used in normalizing flows \cite{papamakarios2021normalizing, ziegler2019latent} or diffusion models \cite{sohl2015deep, ho2020denoising}, and broadens the conventional univariate data generation approach by transposing the cumulative distribution function from a uniform base to preserve the original data's univariate ranks.

The PAI framework is a significant advancement in statistical inference, particularly for unstructured, multimodal, and tabular data. It exceeds traditional methods in reliability and breadth of application, chiefly through synthetic data created by PASS to emulate any statistic's distribution and properties via Monte Carlo testing. This framework, in contrast to classical methods requiring bias corrections, deduces the distribution of a test statistic via an approximated data generation distribution, thereby facilitating finite-sample inference. Additionally, it trumps resampling methodologies by producing independent synthetic samples for inference. This function promotes broader applications, including data integration, sensitivity analysis, and personalization, thereby widening the gamut of statistical inference into new domains. Specifically,  

\begin{enumerate} 
\item[(1)] {\bf Inference for Unstructured and Multimodal Data.} The PAI framework broadens the scope of statistical inference from numerical to unstructured and multimodal data 
through synthetic data generation. Section \ref{sec:deep} demonstrates the validity of PAI 
when PASS estimates the data-generating distribution via pre-trained generative models such as
normalizing flow or diffusion models. 

\item[(2)] {\bf Pivotal Inference.} 
PAI offers exact inference for any pivotal while controlling the Type-I error, 
which surpasses classical methods that necessitate knowledge of a test statistic's distribution, as supported by Theorem \ref{thm2}.

\item[(3)] {\bf General Inference.} The PAI framework enables approximate inference for non-pivotal statistics while maintaining control over Type-I errors. It achieves this by using an estimated distribution well approximating the data-generating distribution, as illustrated in Theorem \ref{thm1}. 

\item[(4)] {\bf Accounting for Modeling Uncertainty.} PAI distinguishes itself from conventional methods by incorporating modeling uncertainty into the Monte Carlo experiments for uncertainty assessment, leading to more credible conclusions.
\end{enumerate}
 
To demonstrate PAI’s capabilities, we address statistical inference challenges in three previously untapped areas: (1) image synthesis, (2) sentiment analysis using DistilBERT \cite{sanh2019distilbert}, and (3) multimodal inference from text to image generation based on text prompts. Moreover, we also contrast PAI with the conformal inference approach \cite{lei2018distribution} for prediction uncertainty in regression problems. In these scenarios, PAI quantifies uncertainty for generative models that involve hyperparameter optimization, considering the statistical uncertainty of such tuning in the inference process and leveraging pre-trained models to refine the accuracy of learning the data-generating distribution. Contemporary research underscores the significance of sample partitioning in inference to avert data dredging \cite{chernozhukov2018double, wasserman2020universal}. Demonstrated through these applications, PAI conducts innovative hypothesis testing for image synthesis, word inference in sentiment analysis, and generated images from varying text prompts via stable diffusion techniques, thus providing uncertainty quantification for numerical and unstructured data where tests are not analytically tractable.

This paper comprises the following sections: Section \ref{sec: PASS-framework} establishes the foundation of PASS, enabling the estimation of any statistic's distribution through Monte Carlo simulations. Section \ref{sec: inference} introduces the PAI framework and the PASS generator. Section \ref{stat} offers a statistical validation of the PAI framework. Section \ref{sec:deep} develops tests for comparing synthetic images generated by diffusion models \cite{sohl2015deep, ho2020denoising} and other deep generative models such as GLOW \cite{kingma2018glow} and DCGAN \cite{radford2015unsupervised}, also addressing the evaluation of word significance in sentiment analysis using DistilBERT,  multimodal inference from texts to images. Section \ref{numerical} presents numerical experiments. This section additionally contrasts the PAI methodology with the conformal inference approach in quantifying prediction uncertainty in regression problems. Supplementary materials include implementation details for the numerical examples, technical specifics, multivariate ranks, and learning theory for normalizing flows.

\section{Perturbation-Assisted Sample Synthesis}\label{sec: PASS-framework}

Given a $d$-dimensional random sample $\Z=(\Z_i)_{i=1}^n$ from a cumulative distribution function (CDF) $F_{\Z}(\cdot)=F_{\Z}(\cdot~;\bm \theta)$, or data-generating
distribution, $\Z_i \sim F_{\Z}$; $i = 1, \ldots, n$, we estimate a statistic $H(\Z)$'s distribution, where 
$\bm \theta$ is a vector of unknown parameters and
$H$ is a vector of known functions that may be analytically intractable. Here,
$\Z$ could be an independently and identically distributed sample or its continuous latent vector
representation obtained through, for example, a latent normalizing flow (\cite{papamakarios2021normalizing,ziegler2019latent}) and VAE \cite{kingma2013auto} for images and a numerical embedding such as BERT-style transformer for texts. Subsequently, we assume that $\Fz$ is absolutely continuous and use the continuous latent vectors of unstructured data or a continuous surrogate of non-continuous data \cite{bi2023distribution}  for a downstream task.

\subsection{Sample Synthesis}

\label{sec:data-p}

{\bf Generation via Transport.} To generate a random sample $\Z^{\prime} = (\Z^\prime_i)_{i = 1}^n$ from a cumulative distribution $F_{\Z}$, we construct a transport $G$ mapping a base distribution of $\U$ to that of $\Z$, preferably simple, like the Uniform or Gaussian, where $\U = (\U_i)_{i = 1}^n$ is a sample from the base distribution $F_{\U}$. In the univariate case, we generate $\Z^{\prime}_i=G(\U_i)$ by choosing $G=F^{-1}_{\Z}$ with $\U_i$ sampled from the Uniform distribution $U[0,1]$; $i = 1, \ldots, n$. However, this generative method is no longer valid in the multivariate case as the multivariate analogy of $F_{\Z}^{-1}$ does not exist. In such a situation, the reconstruction of $G$ mapping $\mathcal R^d$ to $\mathcal R^d$ is challenging.

{\bf Linkage between Generated and Original Data.} Sample $\Z^{\prime}$ generated from the base distribution of $\U$ may not accurately represent $\Z$ if they are unrelated to $\Z$. When $d = 1$, $\Z^{\prime}$ retains the ranks of $\Z$ if $\U$ retains the ranks of $\Z$, by the non-decreasing property of $G = F_{\Z}$. As argued in \cite{bi2023distribution}, $\Z^{\prime}$ connects to the original sample $\Z$ by rank preservation. This aspect is crucial for personalized inference, outlier detection, and data integration. To generalize this concept of rank preservation to the multivariate situation, we consider a transport $T$ mapping from $F_{\Z}$ to $F_{\U}$, which is not necessarily invertible. However, the invertibility ensures a round-trip transformation between $F_{\Z}$ and $F_{\U}$ is uniquely determined. We then align the multivariate ranks of $(\U_i)_{i = 1}^n$ with those of $(T(\Z_i))_{i = 1}^n$, which preserves the ranks of $(\Z_i)_{i = 1}^n$ using its representation $(T(\Z_i))_{i = 1}^n$ in the space of the base variables $\U$. The reader is directed to the supplementary materials for detailed information on multivariate ranks. In other words, this alignment preserves the ranks of $(\Z_i)_{i = 1}^n$ by $(\U_i)_{i = 1}^n$ when $T$ is invertible and recovers the univariate case. In practice, we may reconstruct $G$ with $T = G^{-1}$ as in the case of normalizing flow or treat a non-invertible $T$ separately as in a diffusion model; see subsequent paragraphs for examples.

{\bf Perturbation for Diversity and Protection.} Recent research in denoising diffusion models (\cite{rombach2022high, sohl2015deep, dhariwal2021diffusion}) has demonstrated that adding Gaussian noise in the forward diffusion process and subsequent denoising to estimate the initial distribution $F_{\Z}$ in the reverse process can effectively improve the diversity of generated samples. Moreover, adding noise in a
certain form of data perturbation \cite{shen2022data} can allow $\Z^{\prime}$ to satisfy the differential privacy standard \cite{bi2023distribution} for privacy protection.

\begin{figure}[ht]
    \centering
    \includegraphics[width=\linewidth]{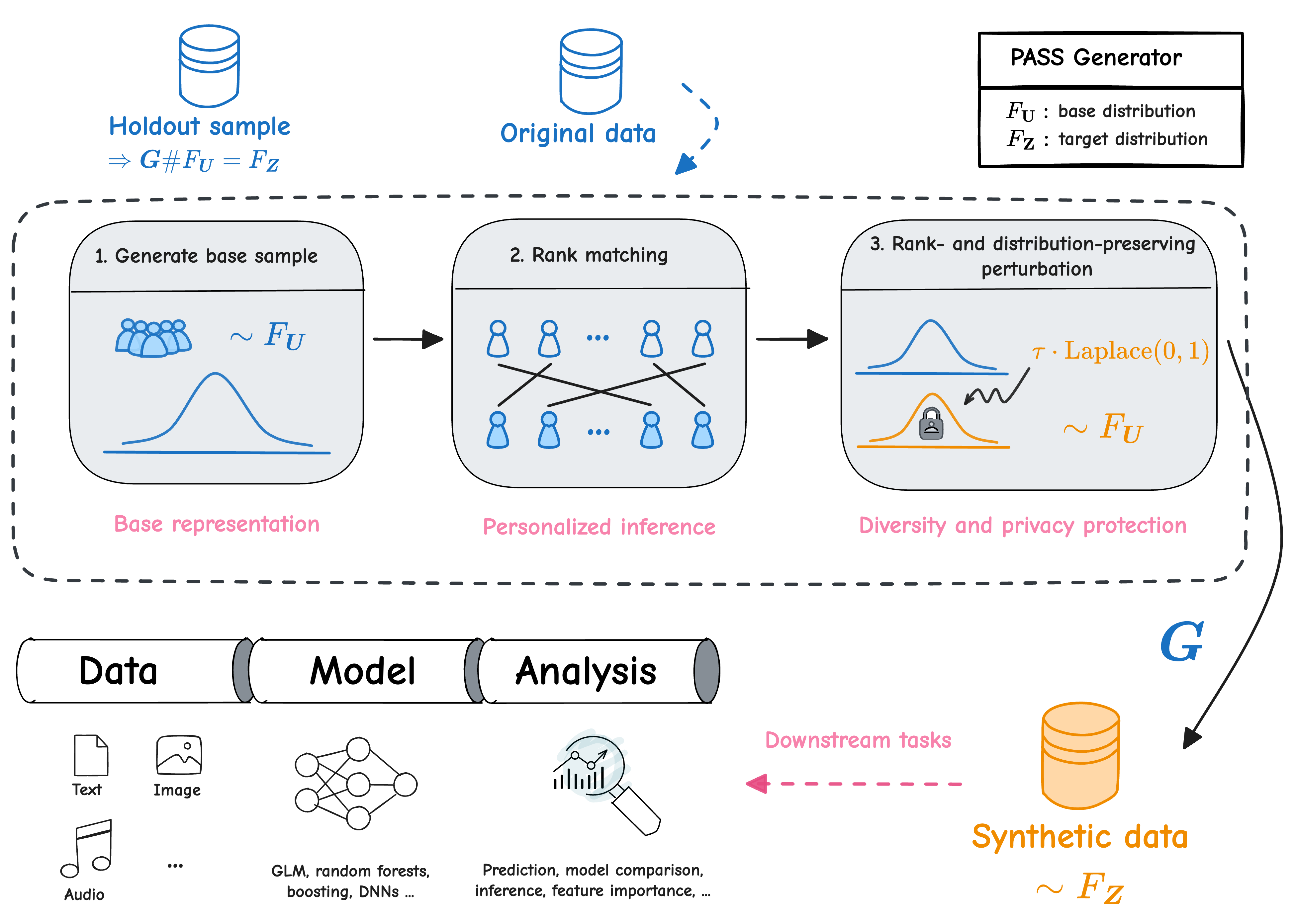}
\caption{Flowchart illustrating the PASS approach with rank matching and distribution-preserving perturbation. PASS generates a synthetic sample that closely retains the multivariate ranks of the original sample, ensuring privacy protection. The transport $G$ is applied to align the base distribution with the target distribution (for example, the original distribution).}
    \label{fig: dpg-flowchart}
\end{figure}

This discussion leads to the generation scheme of PASS, which comprises three components, transport estimation, rank preservation, and data perturbation:

\begin{enumerate}
\item[(1)] Sample $\bm U=(\bm U_i)_{i=1}^n$ from the base distribution $F_{\U}$;

\item[(2)] Compute the permutation map $r(\cdot)$ on $\{1, \ldots, n\}$ to align the multivariate ranks of $(\U_{r(i)})_{i = 1}^n$ with those of $(T(\Z_i))_{i = 1}^n$, where $T$ is a transport
map from $F_{\Z}$ to $F_{\U}$. Please see Section \ref{appendix: multivariate rank} in the supplementary materials for additional details regarding $r(\cdot)$.

\item[(3)] Generate $\Z^{\prime}=(\Z_i^{'})_{i=1}^n$ by adding noise $(\e_i)_{i = 1}^n \sim F_{\e}$: 
\begin{eqnarray} \label{DPG} 
\Z^{\prime}_i = G(\V_i),\quad \V_i=W(\U_{r(i)}; \e_i); i = 1, \ldots, n, 
\end{eqnarray}
where $W$ is a known perturbation function that injects noise to $\bm U$ while preserving the base distribution, that is, $(\V_i)_{i = 1}^n$ will still be random sample from $F_{\V} = F_{\U}$, and $G$ is a transport map that pushes $F_{\V}$ to $F_{\Z}$. An illustration is provided in Fig \ref{fig: dpg-flowchart}.
\end{enumerate}

Notable is that the equation \eqref{DPG} can be applied to embeddings of original data for dimension reduction, as demonstrated in studies such as \cite{rombach2022high, sohl2015deep, dhariwal2021diffusion}. In \eqref{DPG}, $G$ and $T$ represent generation and rank preservation, respectively. For simplicity, we estimate $G$  by assuming 
its inevitability. However, in certain cases, it is advantageous not to impose the invertibility on $G$ while estimating $T$ separately, as in diffusion models. As for perturbation, we can select $W$ to preserve the multivariate ranks of $(\mb U_{r(i)})_{i = 1}^n$ by $\V_i$, even after adding noise $(\mb e_i)_{i = 1}^n$ (see Theorem \ref{theorem: rank preservation}). For example, Section \ref{sec: rank preserve} in the supplementary materials presents an additive form of $W$. 
Regarding the noise distribution $F_{\mb e}$, we typically parametrize it as $\e = \tau \mb \epsilon$, with $\tau > 0$ denoting the perturbation size and $\mb \epsilon \sim F_{\mb \epsilon}$ representing a standardized noise distribution. When privacy is not a concern, we can conveniently set $\tau = 0$ and select $W$ as the identity map. Additionally, when personalization and data integration are not the primary focus, as in Section \ref{sec:deep}, we can choose $r(i) = i$; $i = 1, \ldots, n$.

{\textbf{Separation of $(G,T)$ from a Downstream Task.}}
Ideally, we can repurpose the original sample $\Z$ to estimate the transports $G$ and $T$ while executing a downstream task. However, this approach is debatable regarding the validity of the downstream analysis \cite{devezer2021case}. Whereas it offers valid inference for a pivotal statistic $H(\Z)$, as demonstrated in Theorem \ref{thm2}, it may yield overly optimistic conclusions in post-selection inference \cite{devezer2021case}. To circumvent this problem, we recommend
using an independent holdout sample,  usually available from other studies on the same population. For example, training examples for similar images could serve as holdout data to learn the data-generating distribution for inference, as illustrated in Section \ref{numerical}. By separating downstream analysis from estimating $G$ and $T$, we guarantee the validity of an inference even with finite sample size; see Theorem \ref{thm2}. If holdout data is unavailable, a 
possible alternative is sample splitting, with one subsample acting as a holdout sample. This method can yield valid conclusions but may compromise statistical power \cite{wasserman2009high}.

\subsection{Data-Generating Distribution}

Given a holdout sample $\mathbb{S}_h = (\Z_i)_{i=1}^{n_h}$, our objective is to construct $\tilde F_{\Z}$, or equivalently $\tilde G$, in order to estimate the data-generating distribution $F_{\Z} = F_{\V} \circ G^{-1}$. Building on this foundation, PASS generates $\Z^{\prime} = (\Z^{\prime}_i)_{i=1}^n$ following $\tilde F_{\Z}$, as detailed in Lemma \ref{lem1}. Subsequently, we propose employing generative models to reconstruct $\tilde F_{\Z}$, either \textit{explicitly} by approximating $G$ with an invertible $\tilde G$, as in $\tilde F_{\Z} = F_{\V} \circ \tilde G^{-1}$
as in normalizing flows \cite{kingma2018glow, dinh2014nice, dinh2016density}, or \textit{implicitly} through sampling as in diffusion modeling \cite{sohl2015deep, ho2020denoising}.
Consequently, large pre-trained generative models can enhance the estimation accuracy of the data-generating distribution.

{\bf Explicit Estimation.} We suggest estimating $G$ by maximizing a likelihood function $L(G; \mathbb S_h)$, which is parameterized through the distribution of $\V$. Specifically, we obtain an estimated transport $\tilde G$ by 
\begin{align} \label{mle}
\tilde{G}= \argmin_{G \in \mathcal F} ~ (L(G; \mathbb{S}_h)+\lambda P(G)),
\end{align}
where $\mathcal F$ is a predefined function class, such as normalizing flows, $P(G)$ is a nonnegative penalty function, and $\lambda \geq 0$ is a regularization parameter. In \eqref{mle}, its constrained version can serve the same purpose, as described by \cite{shen2013constrained}. 
Furthermore, due to the nature of $\tilde G$, we can explicitly obtain the analytical form of 
$\tilde F_{\Z}$ and the corresponding density, 
for example, in normalizing flows \cite{kingma2018glow, dinh2014nice, dinh2016density}.

{\textbf{Distribution Estimation of a Statistic $H(\Z)$ by PASS.}}
Given an estimate $\tilde{G}$, we can obtain an estimated distribution $\tilde F_{\Z} = F_{\V}\circ \tilde{G}^{-1}$ when $\tilde G$ is invertible. Notably, PASS can generate synthetic samples using $\Z^{\prime} = \tilde G(\V) \sim \tilde F_{\Z}$ derived from \eqref{DPG}. Based on this, we propose a Monte Carlo method for estimating the CDF $F_{H(\Z)}$ of any statistic $H(\Z)$. Specifically, we generate $D$ independent synthetic samples $(\Z^{\prime(d)})_{d=1}^D$ using \eqref{DPG}, and construct the PASS estimate as an empirical CDF: $\tilde F_{H(\Z^{\prime})}(\bm x) = D^{-1} \sum_{d=1}^D I(H(\Z^{\prime(d)}) \leq \bm x)$ for estimating $F_H$, where each $\Z^{\prime(d)}$ is from $\tilde F_{\Z}$ by PASS. Refer to Section \ref{stat} for statistical guarantee and justification of this approach.

\subsection{Sampling Properties of PASS}
\label{sec: sampling property}

Lemma \ref{lem1} presents the sampling properties of $\Z^{\prime}$ generated by PASS. 

\begin{lemma} (Sampling properties of PASS) \label{lem1}
Given $\Z^{\prime}=(\Z^{\prime}_i)_{i=1}^n$ generated from \eqref{DPG} using $\tilde G$, assume that $\tilde F_{\Z}$ is independent of $\Z = (\Z_i)_{i = 1}^n$. Then, 
\begin{enumerate}
\item (Within-sample) $\Z^{\prime}=(\Z^{\prime}_i)_{i=1}^n$ is an independent and identically distributed 
(iid) sample of size $n$ according to $\tilde F_{\Z}$
when $\Z$ is independent and identically distributed.
\item (Independence) $H(\Z^{\prime})$ is independent of $\Z$ for any
permutation-invariant $H$ in that $H(\Z) = H(\Z_{\pi})$ with $\Z_{\pi} = 
(\Z_{\pi(i)})_{i = 1}^n$, where $\pi$ represents any 
permutation map on $\{1, \ldots, n\}$.
\end{enumerate}
\end{lemma}

Lemma \ref{lem1} highlights the two advantages of a generated PASS sample $\Z^{\prime}$. First, its iid property is unique and not shared by any resampling approach. Second, the conditional distribution of the PASS statistic $H(\Z^{\prime})$ given $\Z$ is the same as its unconditional distribution, a property not shared by existing resampling methods. This aspect is somewhat surprising because the permutation invariance of a test statistic $H$ allows for rank preservation of $\Z^{\prime}$ without imposing dependence between $\Z^{\prime}$ and $\Z$. Note that a common test statistic $H$ is invariant concerning the permutation of the sample order for an iid sample \cite{hollander2013nonparametric}. These two aspects ensure that the PASS sample $H(\Z^{\prime})$ accurately represents $H(\Z)$, leading to a reliable estimate of the distribution of $H(\Z)$.

\section{Perturbation-Assisted Inference}\label{sec: inference}

For inference, data scientists often use a statistic $H(\bm Z)$ for hypothesis testing or a confidence interval concerning  $\bm \theta$ or its functions. Based on the PASS framework described in Section \ref{sec: PASS-framework}, we estimate the distribution of $H(\bm Z)$, which permits a valid inference through Monte Carlo simulation. We introduce a generative inference framework called Perturbation-Assisted Inference (PAI). PAI involves two independent samples: 
an inference sample $\mathbb{S}=(\bm Z_i)_{i=1}^{n}$ via $H(\bm Z)$ and a holdout sample $\mathbb{S}_h=(\bm Z_i)_{i=1}^{n_h}$ for estimating the generating distribution via PASS. However, if $H(\bm Z)$ is pivotal, then holdout and inference samples can be the same, as suggested by Theorem \ref{thm2}.

\begin{figure}[ht]
    \centering
    \includegraphics[width=\linewidth]{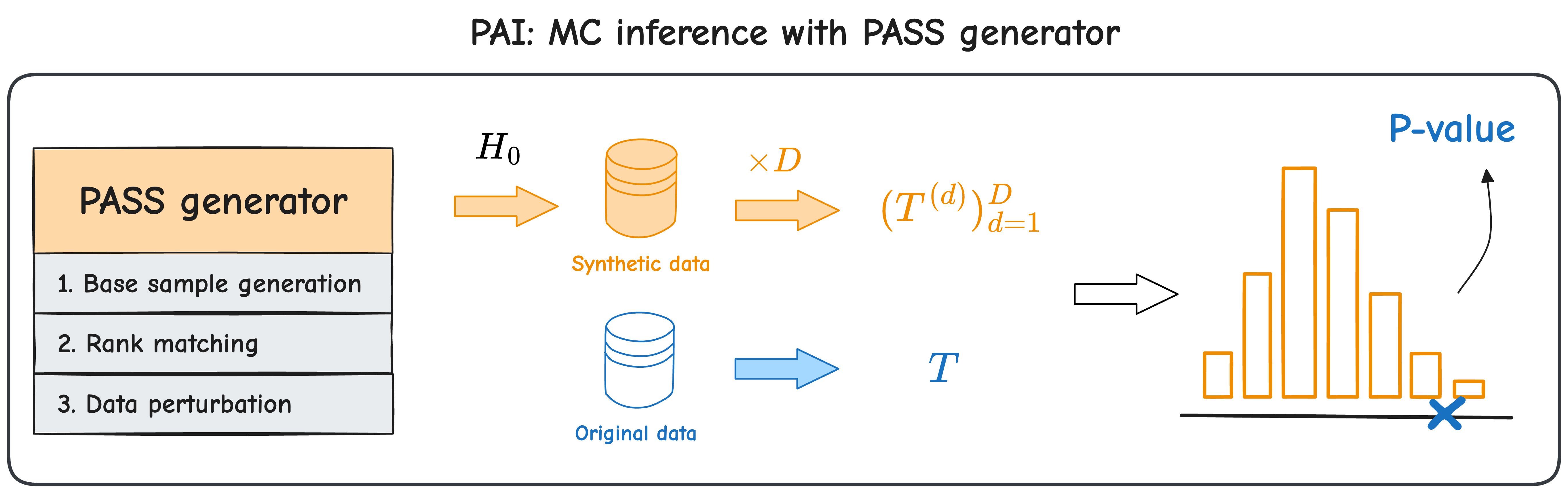}
\caption{Estimating the distribution of the test statistic under the null hypothesis ($H_0$) through Perturbation-Assisted Inference (PAI) using the PASS generator: A Monte Carlo (MC) approach.}
    \label{fig: pai-pass-flowchart}
\end{figure}

To perform a hypothesis test, we proceed as follows:
\begin{enumerate}
    \item[(1)] \textbf{Estimation of Null Distribution of $H(\bm Z)$.} Under the null hypothesis $H_0$, we use the holdout sample $\mathbb{S}_h$ for the 
data-generating distribution of PASS in \eqref{DPG} to estimate the null distribution of a test statistic $H(\bm Z)$, which avoids sample reuse. Specifically, we generate $D$ independent copies of synthetic samples $\Z^{\prime(d)}$; $d=1,\ldots,D$ via \eqref{DPG}, where $\tilde F_{\bm Z}(\cdot)=F_{\bm Z} (\cdot~; \tilde{\bm \theta}^0)$ with $\tilde{\bm \theta}^0$ being an estimate of $\bm \theta$ under $H_0$. Then, we compute the empirical distribution $\tilde F_{H(\Z^{\prime})}(\bm x)= D^{-1} \sum_{d=1}^D I(H(\Z^{\prime(d)}) \leq \bm x)$ for any real $\bm x$ as the PASS estimate of $F_H$ given $D$ independent copies of synthetic samples $\{\Z^{\prime(d)}\}_{d=1}^D$ via \eqref{DPG}, where each sample $\Z^{\prime(d)}$ is from $\tilde F_{\Z}$ by PASS.

    \item[(2)] \textbf{Inference.} We use the empirical null distribution $F_{H(\Z^{\prime})}$ to compute the rejection probability based on a trained machine learner evaluated on an inference sample $\mathbb{S}_i$ to draw an inference. Moreover, we can convert a test into a confidence set.

\end{enumerate}

{\bf Connection with Other Generative Models.} PASS is compatible with various generative models for estimating the transport $G$ in \eqref{DPG}, which can utilize
large pre-trained models to enhance the accuracy of distribution estimation. Unlike other generators, PASS maintains the ranks of an inference sample and incorporates noise to diversify or safeguard the original data.

{\bf Connection with Resampling.} The resampling approach tailors for low-dimensional numerical data \cite{efron1992bootstrap}, where $F_{\Z}$ can be accurately estimated based on $\Z$. However, these methods struggle with high-dimensional data due to the curse of dimensionality. Additionally, the resampled data is only conditionally independent, even when $\Z$ is independent.
For example, in the parametric bootstrap, conditioning on $\Z \sim N(\bm \mu,\bm I)$, a sample $\bm Z^{B} \sim N(\hat{\bm \mu}, \bm I)$ assuming known identity covariance matrix $\mb I$ and $\hat{\mb \mu}$ is the estimated mean vector from $\mb Z$.
However, for its unconditional distribution, $\E \mb Z^B = \E \hat{\bm \mu}$ and $\Var \mb Z^B = \Var \hat{\mb \mu} + \mb I$.
This approach can lead to overly optimistic conclusions in post-selection inference as $\hat{\bm\mu}$ depends on a selected model \cite{faraway1998data,lei2018distribution,wasserman2009high}.

 In contrast, PASS produces an independent sample when a holdout sample is independent of $\Z$, as discussed in Lemma \ref{lem1}, which enables valid inference. Moreover, a PASS sample preserves the ranks of an inference sample, facilitating personalization and data integration. Crucially, PASS can generate numerical, unstructured, and multimodal data, such as image-text pairs, allowing PAI to transcend the traditional inference framework and tackle complex problems involving unstructured and multimodal data inference.

\section{Statistical guarantee and justification}
\label{stat}

Given PASS samples $\Z^{\prime(d)}$ from $\tilde F_{\Z}$
estimated on an independent holdout sample, we provide a guarantee of validity 
of PAI by investigating PASS's estimation error of $\hF_{H(\Z^{\prime})}$, as measured
by the Kolmogorov-Smirnov Distance:
$\text{KS}(\hF_{H(\Z^{\prime})},F_{H})=\sup_{\bm{x}}|\hF_{H(\Z^{\prime})}(\bm{x})-F_{H}(\bm{x})|$. 
Next, we perform the error analysis for non-pivotal inference and pivotal inference.

\subsection{General Inference with Holdout}

\begin{theorem} (Validity of PAI)
\label{thm1} Assume that the estimated data-generating distribution by PASS 
on a holdout sample $\mathbb S_h$ of size $n_h$ is independent of an inference
sample $\mathbb S$. Moreover, $H$ is a permutation-invariant statistic calculated on $\mathbb S$. 
Then, the reconstruction error of 
$\hF_{H(\Z^{\prime})}$ with the MC size $D$ by PASS satisfies: for any small $\delta > 0$, with probability at least $(1 - \delta)$,
\begin{eqnarray}
\label{main:thm:new}
\text{KS}(\hF_{H(\Z^{\prime})},F_{H}) \leq \sqrt{\frac{\log (4/\delta)}{2D}}+ 
|\mathbb S| \cdot \text{TV}(\hFz, \Fz),
\end{eqnarray}
where $\text{TV}(\hFz, \Fz)$ is the total variation distance between the distributions of 
$\hFz$ and $\Fz$. Hence, PAI yields a valid test on $\mathbb S$ provided that 
$|\mathbb S| \cdot \text{TV}(\hFz, \Fz)\rightarrow 0$ as $n_h \rightarrow \infty$ and 
$D \rightarrow \infty$.
\end{theorem}

\begin{remark} Note that $|\mathbb S| \cdot \text{TV}(\hFz, \Fz)\rightarrow 0$ requires that the 
holdout size $n_h=|\mathbb S_h|$ should be larger than the inference size $n=|\mathbb S|$ as
$\text{TV}(\hFz, \Fz) \rightarrow 0$ at a rate slower than $n_h^{-1}$.    
\end{remark}

\begin{remark}
For a diffusion model defined by a $d$-dimensional Brownian motion, Theorem 5.1 of \cite{oko2023diffusion} establishes the error bound between $\hFz$  and $\Fz$:
under regularity conditions: 
$$
\E[\text{TV}(\hFz, \Fz)] =O(n_h^{-r/(2r+d)} (\log n_h)^{\frac{5d+8r}{2d}}),
$$
where the data-generating distribution $\Fz$ belongs to the Besov ball $B^r_{p,q}([-1,1]^d,C)$ 
with radius $C>0$ and the $L_p$-modulus of smoothness $r>d(1/p-1/2)_+$, as
measured by the $L_q$-norm ($p,q>0$). 
\end{remark} 

\begin{remark}
For normalizing flows, Proposition \ref{prop1} in the Supplement Material provides an
error bound for $\text{TV}(\hFz, \Fz)$ expressed in
terms of the estimation and approximation errors of a flow, which implies that 
$\text{TV}(\hFz, \Fz) \rightarrow 0$  as $n_h \rightarrow +\infty$ when the approximation 
error tends to zero, which we expect as a flow serves as a universal approximator for 
complex distributions \cite{koehler2021representational}.
\end{remark}

Theorem \ref{thm1} suggests that the estimation error of the PASS estimate, $\tilde F_{H(\Z^{\prime})}$, is governed by two factors: the Monte Carlo (MC) error, $\sqrt{\frac{\log (4/\delta)}{2D}}$, and the estimation error of the data-generating distribution, $\text{TV}(\tilde{\Z}, \Z)$. The MC error diminishes to $0$ as the MC size, $D$, increases, while the latter depends on the estimation method of $G$ in \eqref{DPG} applied to a holdout sample, $\mathbb{S}_h$, which in general goes to $0$ as $n_h \to +\infty$.
Moreover, PASS can utilize large pre-trained models to boost learning accuracy via knowledge transfer, which we may regard as an increase in $n_h$.


\subsection{Pivotal Inference without Holdout}

This subsection generalizes the previous result to a pivotal $H(\Z)=T(\tilde{\bm \theta},\bm \theta)$ for parameter $\bm \theta$, where $T$ is a transformation and $\tilde{\bm \theta}$ is an estimate of $\bm \theta$ based on $\Z$. In this situation, PAI does not require a holdout sample, $F_{\Z}(\cdot) = F_{\Z}(\cdot~; \bm \theta)$ is parametrized by $\bm \theta$, and $\tilde F_{\Z}(\cdot)=F_{\Z}(\cdot~; \tilde{\bm \theta})=F_{\bm V} \circ G^{-1}(\cdot~; \tilde{\bm \theta})$, where $\tilde{\bm \theta}$ can be any estimator of $\bm \theta$ due to the pivotal property and \eqref{mle} is no longer required.
Moreover, given PASS samples
$\{\Z^{\prime(d)}\}_{d=1}^D$ from $\tilde F_{\Z}$ using PASS,
the PAI pivotal is $H(\Z^{\prime(d)}) = T(\tilde{\bm \theta}^{(d)},\tilde{\bm \theta})$, where
$\tilde{\bm \theta}^{(d)}$ is an estimate of $\bm \theta$ on $\Z^{\prime(d)}$;
$d=1,\ldots,D$.

\begin{theorem} (Validity of PAI for Pivotal Inference)  \label{thm2} 
The conclusion of Theorem \ref{thm1} holds with $\text{TV}(\tilde{\Z}, \Z)=0$ provided that $H(\Z)$ is pivotal for $\bm \theta$. Hence, PAI yields a valid test on $\mathbb S$ provided that
$D \rightarrow \infty$.
\end{theorem}

Theorem \ref{thm2} establishes that the PASS estimate $\tilde F_{H(\Z^{\prime})}$ can exactly recover $F_H$ without any estimation error of the data-generating distribution, provided that $H(\Z)$ is pivotal, even though the estimation error occurs when estimating $F_{\bm Z}$. This result improves the previous findings in \cite{shen2022data} and justifies using an inference sample $\mathbb S$ alone to estimate
$F_{\Z}$ for making pivotal inferences. 

\section{Applications} \label{sec:deep}

\subsection{Image Synthesis} \label{sec:images}

In image synthesis, deep generative models have been popular due 
to the quality of generated synthetic images from original images. Recently, 
researchers have demonstrated that cascaded diffusion models \cite{ho2022cascaded} can generate high resolution with high-fidelity images surpassing BigGan-deep 
\cite{brock2018large} and VQ-VAE2 \cite{razavi2019generating} on the Fr\'echet inception distance (FID). However, such a comparison lacks uncertainty quantification. Subsequently, we fill the gap to draw a formal inference with the uncertainty quantification
for comparing two distributions. 

Given two multivariate Gaussian distributions $P_0 = N(\bm \mu_0, \bm \Sigma_0)$ and $P = N(\bm \mu, \bm \Sigma)$, the FID score is defined as $\text{FID}(P_0, P) = \|\bm \mu_0 - \bm \mu\|_2^2 + \text{tr}\paren{\bm 
\Sigma_0 + \bm \Sigma - 2\paren{\bm \Sigma \bm \Sigma_0}^{\frac{1}{2}}}$, where $\|\cdot\|_2$ is the $L_2$-norm, and $\text{tr}$ denotes the trace of a matrix.
For measuring the quality of generated images, we usually calculate FID on the feature maps extracted via Inception-V3 model \cite{szegedy2017inception}, a pre-trained vision model that has a great capacity for extracting visual signals.
In our case, $P_0$ and $P$ would be the original and generated distributions of those feature maps.
Here, we test   
\begin{eqnarray}
\label{hypothesis1}
H_0: \text{FID}(P_0, P) = 0, \quad H_a: \text{FID}(P_0, P) > 0.
\end{eqnarray}
Then, we construct a test statistic as follows: 
$T=\text{FID}(\hat P_0, \hat P)$, the empirical FID score between the empirical distribution 
of test images $\hat P_0$ and that of synthesized test images $\hat P$ using a 
trained model, on feature maps from the Inception-V3 model.

\begin{figure}[ht]
    \centering
    \includegraphics[width=\linewidth]{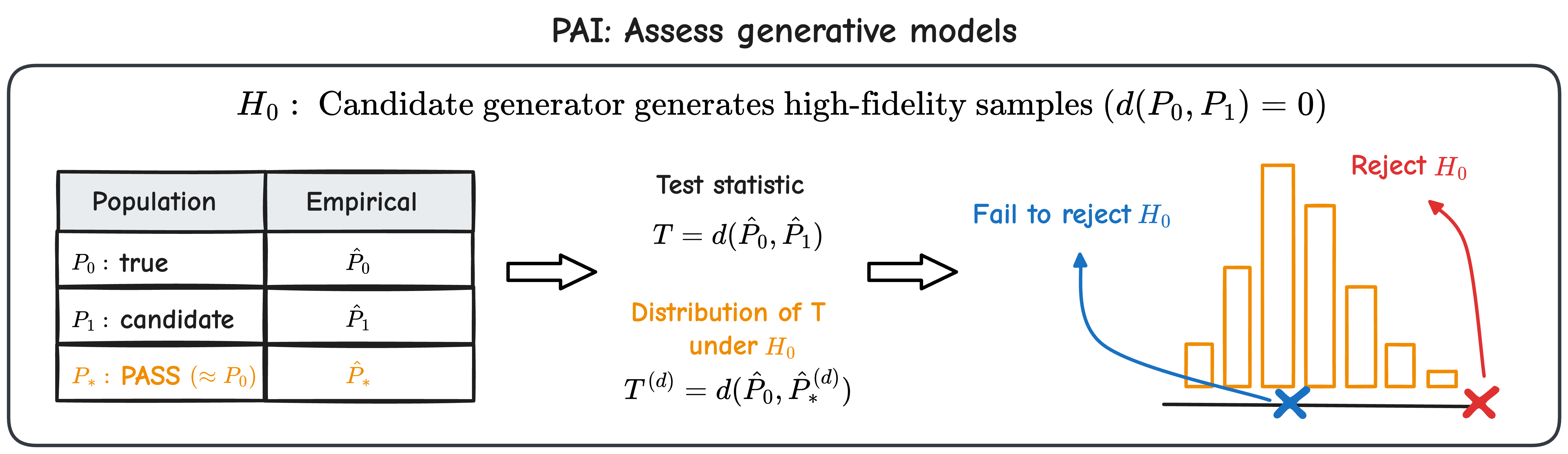}
\caption{Illustration of assessing generative models using PAI. $d(\cdot, \cdot)$ represents distributional distance. A test statistic in the tails (red) suggests statistical evidence against the candidate model generating high-fidelity samples. Conversely, a test statistic near the mode (blue) indicates the opposite. For further details, see Algorithm \ref{algorithm: eg1} in the supplementary materials.}
    \label{fig: pai-assessment}
\end{figure}

To train PASS for PAI inference, we create two independent sets of images denoted by $\mathbb{S}_h 
= (\Z_i)_{i=1}^{n_h}$ and $\mathbb{S} = (\Z_i)_{i=n_h + 1}^{n_h + n}$ for holdout and inference, where $\Z_i$ represents the $i$-th image. For image generation, we further split the inference sample $\mathbb{S}$ into training and test sets for training and evaluating a generator, which is a common practice. Then, we proceed in three steps.
First, we train a PASS generator on a holdout sample $\mathbb{S}_h$ to generate the null distribution under the null that there is no difference between the PASS and the candidate generators under $H_0$. Second, we train a candidate generator on the training set, with which we evaluate its performance using the test statistic $T = \text{FID}(\hat{P}_0, \hat{P})$ on the test set, 
where $\hat{P}_0$ and $\hat{P}$ are the estimated distributions from the baseline and the candidate generator.
Third, we generate $D$ independent copies of synthetic images $(\Z_i^{\prime(d)})_{i=1}^n$ from the null distribution using PASS; $d = 1, \ldots, D$.
Then, we compute the corresponding test statistics $(T^{(d)})_{d=1}^D$ to
obtain the empirical null distribution of $T$ on $\mathbb{S}$, where $T^{(d)}=\text{FID}(\hat{P}_0, \hat{P}^{(d)})$ evaluated on $\mathbb{S}$, and $\hat{P}^{(d)}$ is obtained on $(\bm{Z}_i^{\prime(d)})_{i=1}^n$. Finally, we compute a two-sided\footnote{Given that the knowledge
is unknown concerning the performance of a candidate generator over the PASS generator, we perform a two-sided test to avoid Type-III error.} P-value using $(T^{(d)})_{d=1}^D$ and $T$ based on $\mathbb{S}$. For detailed steps of this computation, refer to Algorithm \ref{algorithm: eg1} in the supplementary materials. An illustrative representation of this procedure can be found in Figure \ref{fig: pai-assessment}.

\subsection{Sentiment Word Inference} \label{nlp}

Given the unstructured nature of data and the complexity of modeling techniques such as transformer-based models, inferring important words for a learning task can be challenging.
In this section, we perform a significance test for the feature relevance of a collection of
positive, negative, and neutral words for sentiment analysis of text reviews labeled as positive or negative.

Let $\mc W_M$ be the words of interest. Consider the null hypothesis $H_0$ and its alternative hypothesis $H_a$:
{\small 
\begin{eqnarray}
\label{hypothesis2}
H_0: R(f^0) - R(f^0_{\mc W_M}) = 0, \quad H_a: R(f^0) - R(f^0_{\mc W_M}) < 0,
\end{eqnarray}
}
where $R$ represents the risk under the data distribution, and $f^0$ and $f^0_{\mc W_M}$ are two population risk minimizers of decision functions on all words $\mc W$ and those with masked words of $\mc W$, respectively. The masked words of $\mc W$ are those highly attended words of $\mc W$ by transformer-based models such as BERT \cite{devlin2018bert} on training samples. It is important to note that masking highly attended words of $\mc W$ is crucial since state-of-the-art embedding models such as BERT can infer words that other embedding models such as Word2Vec \cite{mikolov2013efficient} are incapable of. For more details, refer to Section \ref{ex:nlp}.

\begin{figure}[ht]
    \centering
    \includegraphics[width=\linewidth]{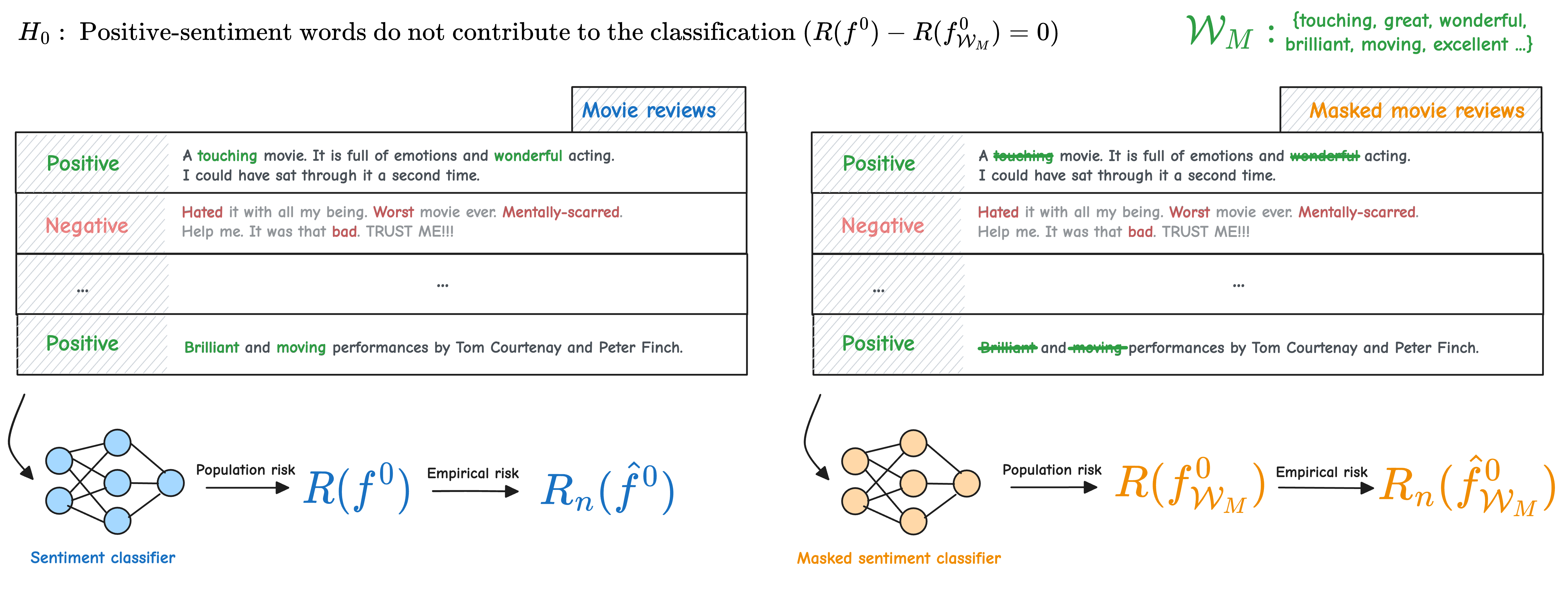}
\caption{Illustration of the black-box test statistic \cite{dai2024significance} employed for assessing feature significance within sentiment classification. If the tested words hold importance for the classification, the risk associated with the masked classifier is expected to be elevated.}
    \label{fig: sentiment_test_stat}
\end{figure}

PAI constructs a test statistic $T$ using the empirical risk $R_n$ evaluated on an inference sample that is independent of the training sample:
\begin{eqnarray}
\label{tests}
T = \frac{R_n(\hat{f}) - R_n(\hat{f}_{\mc W_M})}{SE(R_n(\hat{f}) - R_n(\hat{f}_{\mc W_M}))} ,
\end{eqnarray}
where $\hat{f}$ and $\hat{f}_{\mc W_M}$ are the corresponding trained decision functions, $R_n$ is the empirical risk evaluated on an independent inference sample, and $SE(\cdot)$ denotes the standard error. Refer to Figure \ref{fig: sentiment_test_stat} for a visual representation of the test statistic.

For sentiment analysis, we further split the inference sample $\mathbb{S}$ into training and test sets for training and evaluating a classifier, as in Section \ref{sec:images}. Then, we proceed in three steps.
First, we train normalizing flows on $\mathbb{S}_h$ to generate the joint null distribution of masked embeddings and sentiment labels under $H_0$.
Second, we train sentiment classifiers $\hat{f}$ and $\hat{f}_{\mc W_M}$ respectively on the training set and its masked version to get test statistic \eqref{tests}.
Third, we generate $D$ datasets on the embedding space $\mathbb{E}^{(d)}=(\X_i^{\prime(d)}, Y_i^{\prime(d)})_{i=1}^n$ from the null distribution estimated by PASS to compute corresponding test statistics $(T^{(d)})_{d=1}^D$ on the test set to obtain the empirical null distribution of $T$, where $\mb X_i^{\prime(d)}$ and $Y_i^{\prime(d)}$ represent embedding and corresponding sentiment label, $T^{(d)} = \bar R^{(d)} / SE(\bar R^{(d)})$ and $\bar R^{(d)} = R_n(\hat{f}^{(d)}) - R_n(\hat{f}^{(d)}_{\mc W_M})$ are calculated on $\mathbb{S}^{(d)}$.
Finally, we obtain the P-value of $T$ evaluated on the test set by comparing its value with the empirical distribution of $(T^{(d)})_{d = 1}^D$, c.f., Algorithm \ref{algorithm: eg2} in the supplementary materials for details and Figure \ref{fig: pai_sentiment} for an illustration of this procedure.

\begin{figure}[ht]
    \centering
    \includegraphics[width=\linewidth]{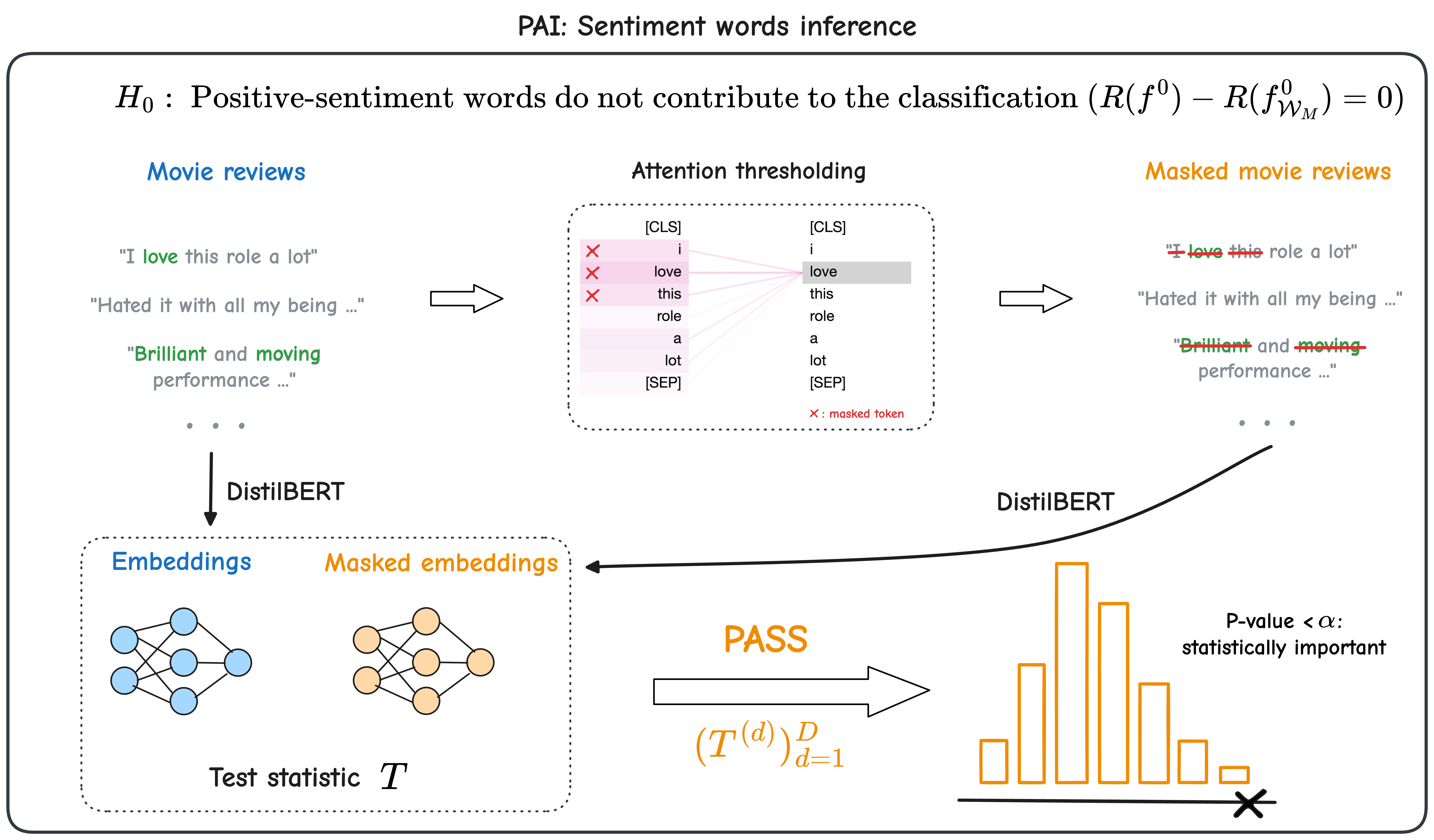}
\caption{Depiction of sentiment words inference using PAI. Words under test and their contextual surroundings are masked according to attention thresholds to compute the test statistic; detailed explanation in Section \ref{ex:nlp}. PAI operates within the embedding space formulated by DistilBERT; see Algorithm \ref{algorithm: eg2} in the supplementary materials for comprehensive steps.}
    \label{fig: pai_sentiment}
\end{figure}

\subsection{Text-to-Image Generation}
Stable Diffusion, a latent diffusion model \cite{rombach2022high}, can generate detailed images given a text prompt. This subsection performs a conditional inference to quantify the statistical certainty of text-to-image generation. Given two text prompts $\x^{(1)}$ and $\x^{(2)}$, we construct a 
coherence test for corresponding generated images $\Y^{(1)}$ and $\Y^{(2)}$ by contrasting their conditional distributions $P(\mb y|\x^{(1)})$ and $P(\mb y|\x^{(2)})$.

For uncertainty quantification, we use the Inception-V3 embeddings $\mathbf{e}^{(k)}$ \cite{szegedy2016rethinking} for images $\Y^{(k)}$; $k = 1, 2$. Under the Gaussian assumption \cite{heusel2017gans}, we define the $\text{FID}$ score $\text{FID}(P_1, P_2$ between the distributions of two embeddings $\e^{(k)}$; $k=1,2$,
as a coherence measure for hypothesis testing:
{\small 
\begin{equation}
\label{hypothesis: multimodal}
H_0: \text{FID}(P_1, P_2) = 0, \quad H_a: \text{FID}(P_1, P_2) > 0.
\end{equation}}
Moreover, we construct $T = \text{FID}(\hat P_1, \hat P_2)$ as a test statistic, where 
$\hat P_k$ is the corresponding empirical distribution of image embeddings on an inference sample of size $n_k$; $k = 1, 2$. 

For PAI inference, we use the pre-trained Stable Diffusion model \cite{rombach2022high}, a state-of-the-art text-to-image generative model, as our PASS generator.
Then, we apply PASS to simulate the null distribution of test statistic $T$.
Given prompt $\mathbf{x}^{(k)}$, for $d = 1, \dots, D$, PASS generates synthetic samples from $P_k$, resulting in synthetic embeddings $(\e_i^{(k)})_{i=1}^{n_1+n_2}$, of which $(\e_i^{(k)})_{i=1}^{n_1}$ and $(\e_i^{(k)})_{i=n_1 + 1}^{n_1 + n_2}$ are used to calculate FID score $T^{(d)}_k$, which then renders a sample of the test statistic $(T^{(k)}_d)_{k=1,2; d=1,\ldots, D}$ of size $2D$, under the null hypothesis.
Under the null that $P_1 = P_2$, there is no difference between the distribution of $\mb e_i^{(1)}$ and that of $\mb e_j^{(2)}$, and thus 
$T^{(d)}_k$ would be a good estimate for the FID score under $H_0$, using synthetic samples from PASS. Additionally, 
$T^{(d)}_k$ also accounts for the symmetry between $P_1$ and $P_2$ when calculating FID score.
By randomly mixing them, we obtain an estimated null distribution for $T$; 
c.f., Algorithm \ref{algorithm: eg3} in the supplementary materials for details and Figure \ref{fig: pai_multimodal} for an illustration of this procedure.

\begin{figure}[ht]
    \centering
    \includegraphics[width=\linewidth]{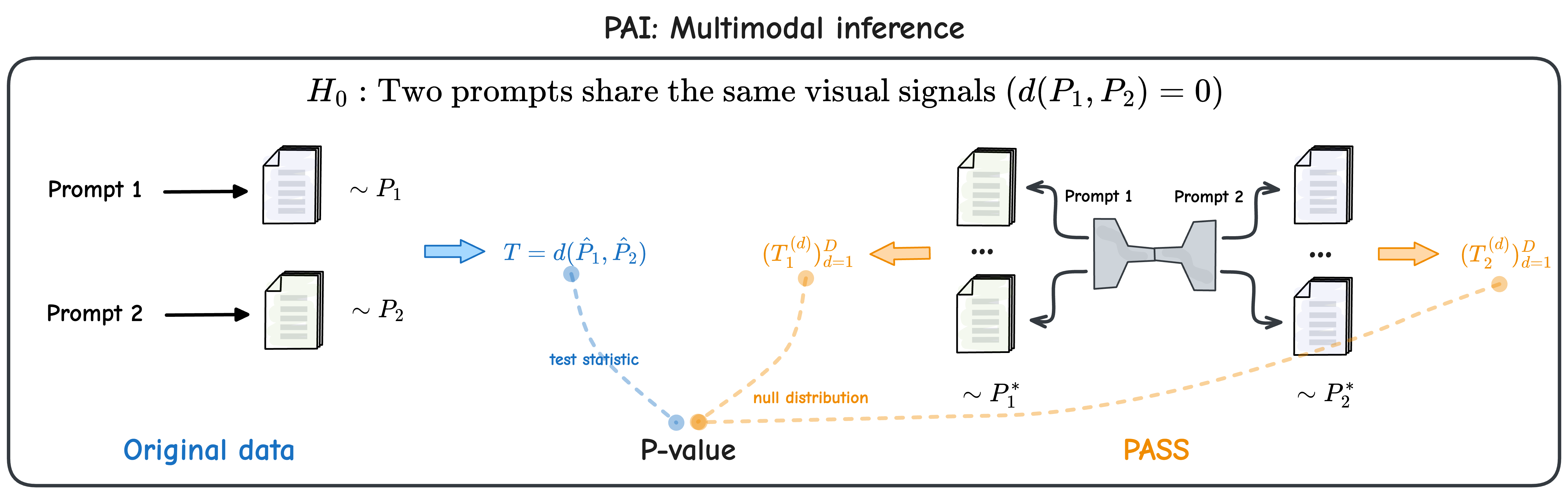}
\caption{Illustration of performing multimodal inference using PAI. Simulated test statistics from both prompts using PASS under $H_0$ are mixed to obtain the estimated null distribution; See Algorithm \ref{algorithm: eg3} in the supplementary materials for details.}
    \label{fig: pai_multimodal}
\end{figure}


\section{Numerical Results}
\label{numerical}

\subsection{Image synthesis}
\label{ex:images}

This subsection applies PAI in Section \ref{sec:deep} to hypothesis testing \eqref{hypothesis1} on the quality of image synthesis using the CIFAR-10 benchmark \cite{krizhevsky2009learning}. This dataset consists of $60,000$ images of size $(3 \times 32 \times 32)$ in 10 different classes, with $50,000$
training and $10,000$ and test images.

\begin{figure}[ht]
    \centering
    \includegraphics[width=\linewidth]{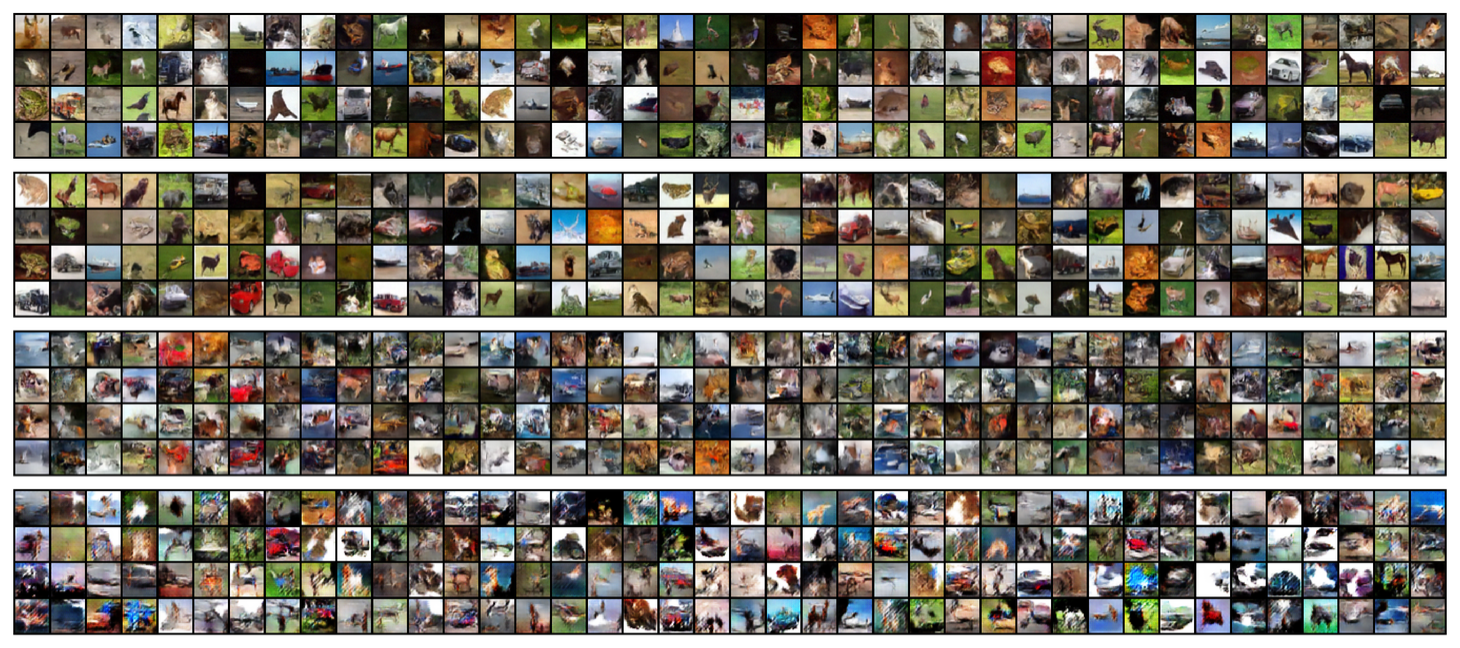}
    \caption{Generated CIFAR-10 images with dimensions (3, 32, 32), using PASS, DDPM, GLOW, and DCGAN methods (from top to bottom), trained on a dataset of 25,000 images.}
    \label{fig: compare-imgs}
\end{figure}

To synthesize images, we use the CIFAR-10 training set while we use a randomly selected subset
of size $n$ of the CIFAR-10 test set for inference.  Additionally, we split the CIFAR-10 training 
set equally into a holdout sample of size $n_h=25,000$ and another sample of size $n_t=25,000$,
respectively for training a PASS generator (reference) and training competitor generators.
In \eqref{DPG}, we use a diffusion model (DDPM, \cite{ho2020denoising}) as our baseline generator,
denoted by PASS-DDPM. We compare the FID scores of three candidate generators against 
the baseline generator PASS-DDPM, including DDPM, deep convolutional GAN (DCGAN, \cite{radford2015unsupervised}), and generative flow (GLOW, \cite{kingma2018glow}); see Fig \ref{fig: compare-imgs} 
for  samples of the generated images by these generators. 
To compute the FID scores, we use a 2048-dimensional feature map extracted from an intermediate layer of a pre-trained Inception-V3 model \cite{szegedy2016rethinking} 
on generated images.

\begin{figure}[ht]
    \centering
    \includegraphics[scale=0.23]{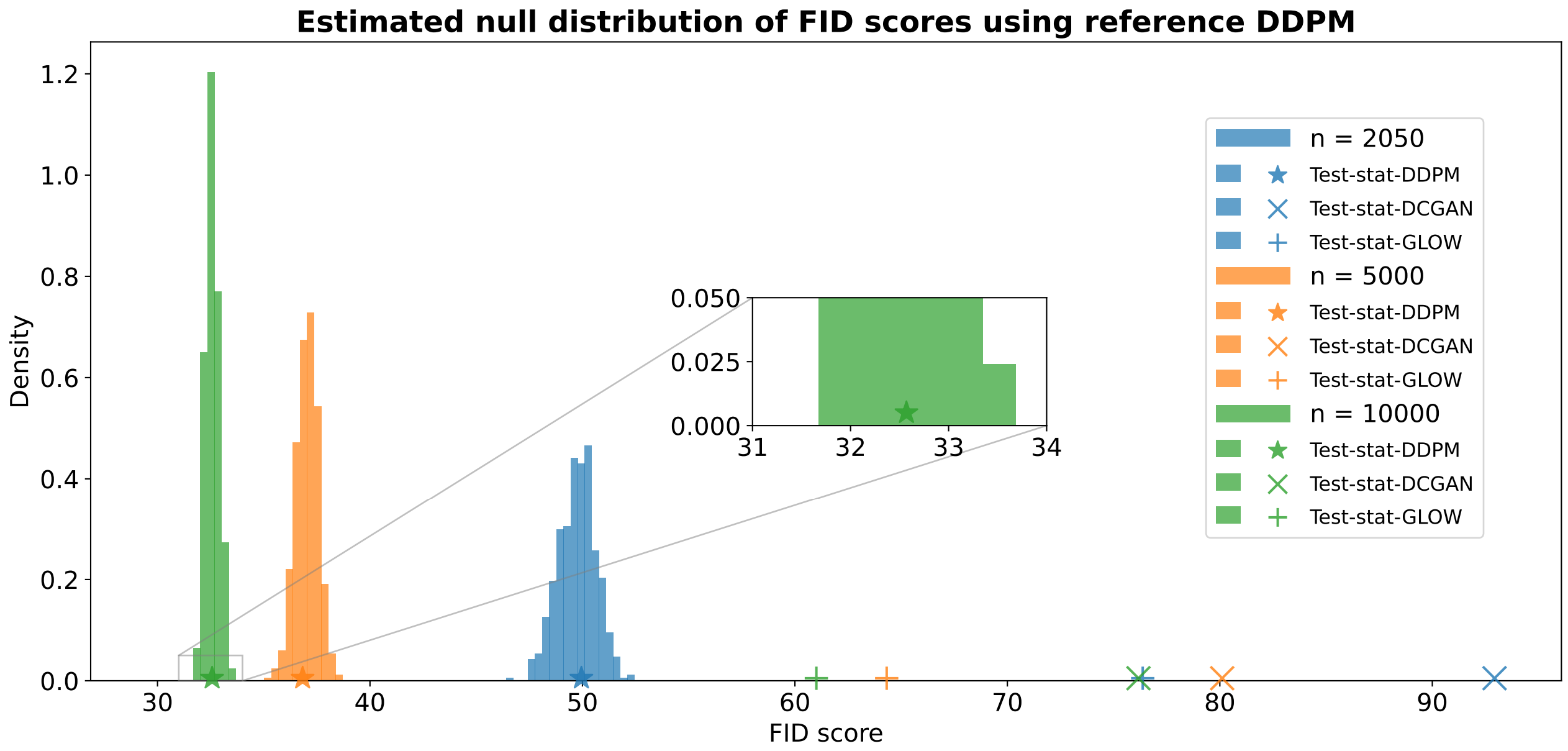}
    \caption{Empirical FID score distributions with inference sizes                  
$n=2050,5000,10,000$ based on $D = 500$ PASS samples from our PASS-DDPM, represented
by colors blue, orange, and green. The FID score is computed using
2048-dimensional features extracted from the Inception-V3 model \cite{szegedy2016rethinking}.}
    \label{fig:dist-fid}
\end{figure}

For the hypothesis test in \eqref{hypothesis2}, we use PASS-DDPM with $D=500$ to estimate the null distribution of the FID score and then compute the corresponding P-value for an inference sample, as shown in Table \ref{tab:test-fid-score}. Fig \ref{fig:dist-fid} illustrates that the empirical null distribution of the FID score varies with the inference sample size $n$ and becomes more concentrated as $n$ increases. This observation highlights the importance of performing uncertainty quantification for the FID score since relying solely on the numerical score could be misleading. Furthermore, we find that DDPM, a generator similar to PASS-DDPM, has a P-value of $.78$, indicating no significant deviation from the baseline PASS-DDPM. However, DCGAN and GLOW exhibit substantial differences from PASS-DDPM, with P-values of $.00$ at a significance level of $\alpha=.05$. We confirm this conclusion as the inference sample size increases from $n=2,050$ to $n=10,000$.

\begin{table}[!ht]
\centering
\caption{FID scores and their P-values for testing \eqref{hypothesis1}, comparing
three generators, DDPM, DCGAN, and GLOW, against the baseline PASS-DDPM. Here
FID scores are computed on 2048-dimensional feature maps of the Inception-V3 model \cite{szegedy2016rethinking} 
with $n$ test and $n$ synthesized images, and DIST-AVG denotes the average FID scores of PASS-DDPM.}
\scalebox{.8}{
\begin{tabular}{@{}cccccc@{}}
\hline
Inf size/Generator     &         & DDPM  & DCGAN & GLOW   & DIST-AVG \\ \hline
\multirow{2}{*}{$n = 2,050$}  & FID     & 49.55 & 92.93 & 76.37 & 49.75     \\
                              & P-value & ~.78  & ~.00  & ~.00   &           \\ \hline
\multirow{2}{*}{$n = 5,000$}  & FID     & 36.83 & 80.11 & 64.32 & 37.04     \\
                              & P-value & ~.65  & ~.00  & ~.00   &           \\ \hline
\multirow{2}{*}{$n = 10,000$} & FID     & 32.57 & 76.17 & 61.01  & 32.58     \\
                              & P-value & ~.72  & ~.00  & ~.00   &           \\ \hline
\end{tabular}}
\label{tab:test-fid-score}
\end{table}

The experiment result shows that DDPM generators are comparable to the baseline PASS-DDPM, but DCGAN and GLOW show significant differences. It underscores the need to account for uncertainty in the FID score to avoid drawing incorrect conclusions about the generation performance.

\subsection{Sentiment Word Inference}
\label{ex:nlp}

This subsection applies PAI to construct a significance test for quantifying the relevance of sentiment collections of positive, negative, and neutral words, in the context of sentiment classification on the IMDB benchmark \cite{maas2011learning}. This dataset comprises 50,000 movie reviews labeled as positive or negative. The goal is to determine whether each collection of words contributes significantly to sentiment analysis.

To perform sentiment analysis, we use a pre-trained DistilBERT model \cite{sanh2019distilbert} to
generate text embeddings. Then, we estimate the null distribution of a test statistic using a normalizing 
flow with a holdout sample of size $n_h=35,000$, followed by the test \eqref{hypothesis1} in Section \ref{nlp} with an inference sample of size $n=5,000$ with a sentiment classifier trained on an independent training set 
of size $10,000$. 

{\bf Extraction of Sentiment Words.} We extract positive and negative sentiment words of IMDB reviews while treating any remaining words as neutral words based on the opinion lexicon provided by \cite{hu2004mining}. Then, we 
extract $|\mb W_M| = 600$ most frequent positive and negative, and neutral words in
each collection for inference. Table \ref{tab:word-inference} displays subsets of 
these words. 

{\bf Masking Contexts of Sentiment Words.}  One main challenge is that
BERT-like models \cite{devlin2018bert} have the capability of inferring
the context information of sentences via unmarked words 
due to the use of masked-language modeling for training. As a result, simply 
masking uni-gram sentiment words does not impact sentiment analysis. To solve this issue, we propose to mask the context of each target word by thresholding attention weights from a pre-trained transformer encoder such that 2\% of the context words are masked.

{\bf Training via Transfer Learning.} To perform sentiment analysis, we construct a classifier by appending a classification head to a pre-trained uncased base DistilBERT model \cite{sanh2019distilbert}, a lighter version of BERT, which permits efficient comprehending of the context. We then fine-tune the model using IMDB review data and obtain fine-tuned embeddings for subsequent tasks. As a result, the model achieves high test accuracy with only a few epochs of fine-tuning.

{\bf Learning Embedding Distribution by Normalizing Flows.}
To train a PASS on the embedding space, we train a RealNVP \cite{dinh2016density} with affine coupling layer on an independent holdout sample $n_h=35,000$ to learn the joint distribution of the pair of text embedding and sentiment 
label. Specifically, we first learn the marginal distribution of sentiment labels and then use normalizing flows to learn the conditional distribution of text embeddings given a sentiment label. The learned flow will be used to emulate the null distribution of the test statistic.
For more training details, please refer to Section \ref{training: eg2} in the supplementary materials.
As Fig \ref{fig: distilbert embeddings} suggests, PASS produces an accurate joint null distribution of the word-label pair, evident from the corresponding marginal and conditional distributions given the label.

{\bf PAI.} We apply PASS to generate $D = 200$ synthetic samples from the null distribution learned from the normalizing flows. Then, we use a training sample of
size $n_t=10,000$ to train a classification model while computing the test statistic on 
the inference sample of size $n=5,000$, with the same splitting ratio for all 
synthetic samples.

\begin{table}[!ht]
\centering 
\caption{Degree of the importance of three collections of 600 of positive, negative, neutral words, 
as measured by the P-value against the irrelevance of each collection by PAI with an MC size $D = 200$.}
\scalebox{.63}{
\begin{tabular}{@{}llc@{}}
\toprule
\textbf{} & \multicolumn{1}{c}{Selected sentiment words} & P-value \\ \midrule
Positive &
  \begin{tabular}[c]{@{}l@{}}\textit{``gratitude, radiant, timely, robust, optimal, thoughtfully, cooperative,}\\ \textit{calming, assurance, oasis, elegant, remarkable, restored, fantastic,}\\ \textit{diplomatic, fastest, excellence, precise, brisk, warmly, ..."}\end{tabular} &
  ~.045 \\ \midrule
Negative &                    
  \begin{tabular}[c]{@{}l@{}}\textit{``cringed, vomit, excuse, vomiting, fails, ashamed, boring, limp,}\\\textit{ ridiculous, aground, scrambled, useless, snarl, annoying, bland,}\\\textit{ unnatural, incorrectly, dire, idiot, leaking, ..."}\end{tabular} &
  ~.015 \\ \midrule
Neutral &
  \begin{tabular}[c]{@{}l@{}}\textit{``administering, reorganized, curving, gleamed, relinquished, circled,}\\ \textit{seeded, streamed, curved, scholastic, canning, accommodated, voluntary,}\\\textit{ cooled, rained, defected, regulated, ousted, straightening, renaming, ..."}\end{tabular} &
  ~.715 \\ \bottomrule
\end{tabular}
\label{tab:word-inference}}
\end{table}

\begin{figure}[ht]
    \centering
    \includegraphics[scale=0.29]{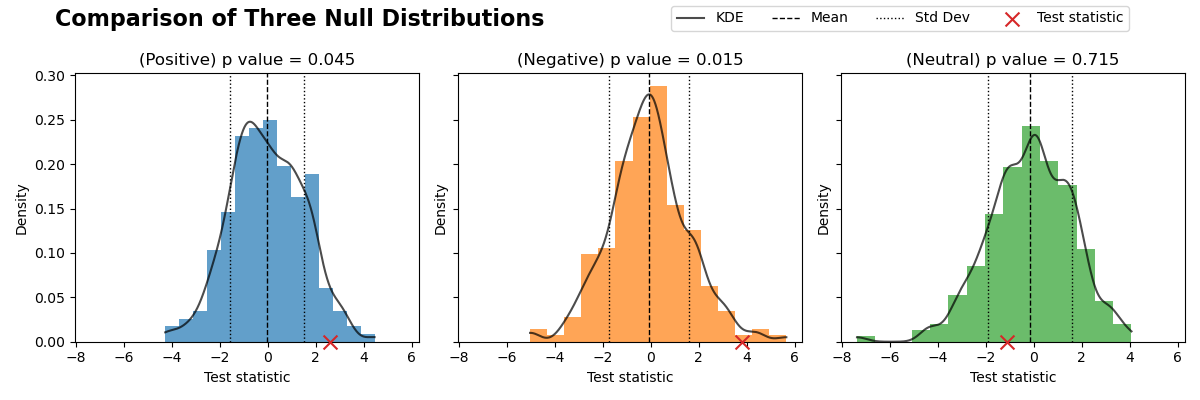}
\caption{Empirical null and their kernel smoothed distributions of the test statistic $T$ for positive (blue), negative (orange), and neutral (green) sentiment words, based on PASS with an MC size $D = 200$ for the hypothesis \eqref{hypothesis2}. Here, red crosses represent the test statistic's values calculated on an inference sample, while the dashed line and two dotted lines represent the empirical mean and standard error.}
    \label{fig: null-dist}
\end{figure}

Table \ref{tab:word-inference} and Fig \ref{fig: null-dist} show that positive and negative words have significant P-values of $.045$ and $.015$, while neutral words are insignificant with a P-value of $.715$, at a significance level of $\alpha =.05$. In other words, positive and negative sentiment words, particularly their contexts, are important predictive features for sentiment analysis.

\begin{figure}[ht]
    \centering
    \includegraphics[scale=0.07]{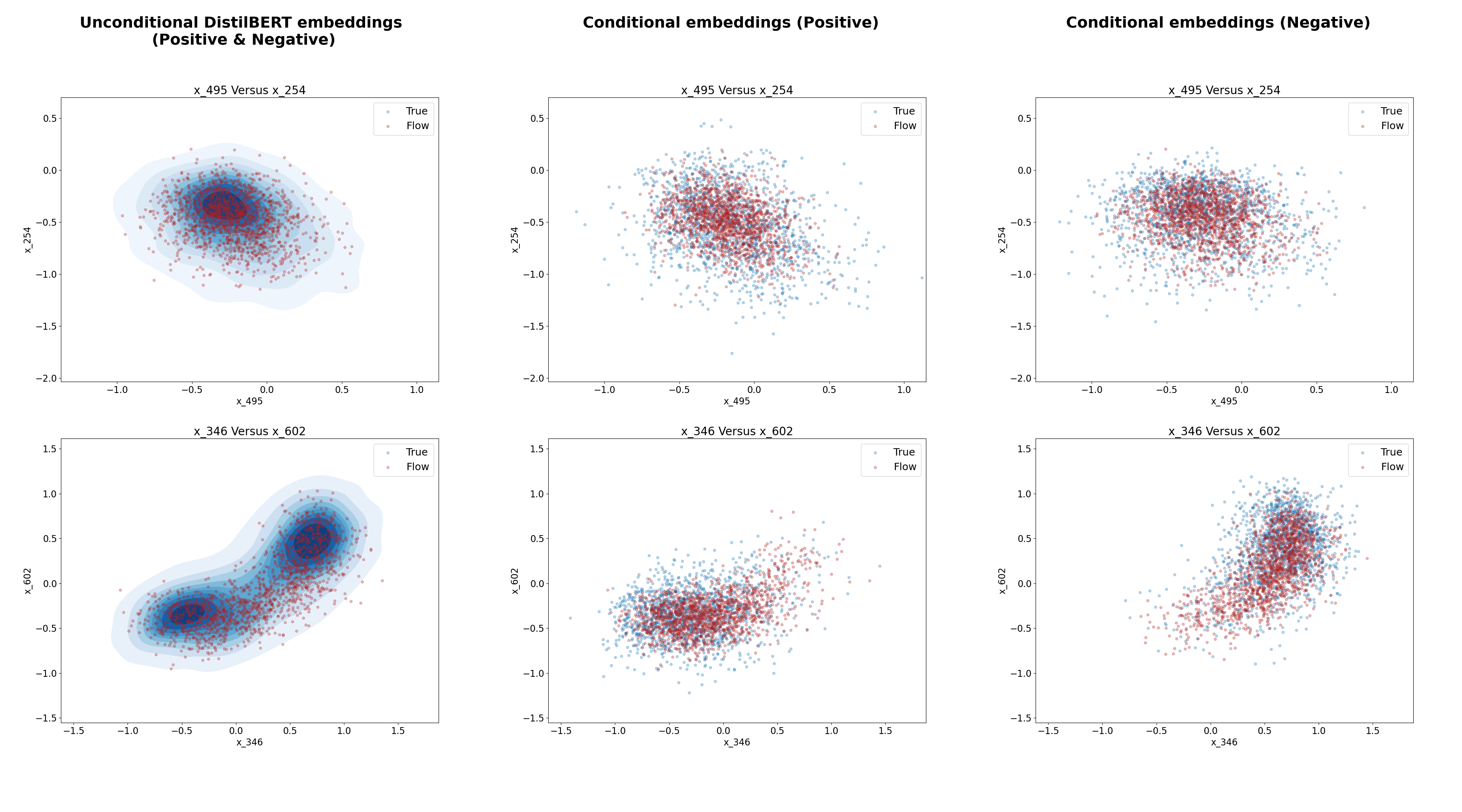}
\caption{Two-dimensional projections from null distributions by PASS (blue) from an
affine coupling flows trained on a holdout sample of size $n_h=35,000$ versus the true distribution (red) via 768-dimensional DistilBERT embeddings. The marginal distribution of combined words and conditional distributions for positive and negative reviews are from left to right. }
    \label{fig: distilbert embeddings}
\end{figure}

To understand the contribution of PASS for simulating the test statistic null distribution, we note that the joint null distribution of positive, neutral, or negative words does not follow the standard Gaussian with an MC size of $D=200$, as indicated by Table \ref{tab:ks-test}. Their distributions differ significantly from their asymptotic distributions \cite{dai2024significance}, despite their smooth curves resembling the Gaussian distribution, as shown in Fig \ref{fig: null-dist}. As a result,   the asymptotic test in \cite{dai2024significance} is not appropriate in this situation. This result demonstrates the usefulness of PASS when a test statistic's distribution significantly deviates from its asymptotic distribution.

\begin{table}[!ht] 
\caption{The Kolmogorov-Smirnov test for the discrepancy between
the test statistic's distribution and the standard Gaussian.}
\centering 
\scalebox{.75}{
\begin{tabular}{@{}lcccc@{}}
\toprule
         & Empirical mean & Std Err & KS test statistic & P-value (two-sided) \\ \midrule
Positive & -.028         & 1.530   & .124             & .004                            \\
Negative & -.083         & 1.659   & .128             & .002                            \\ 
Neutral  & -.137         & 1.749   & .157             & .000                           \\ \bottomrule 
\end{tabular}
\label{tab:ks-test}} 
\end{table}

\subsection{Text-to-Image Generation}

\label{tab:multimodal}

\begin{figure}[ht]
    \centering
    \includegraphics[scale=0.226]{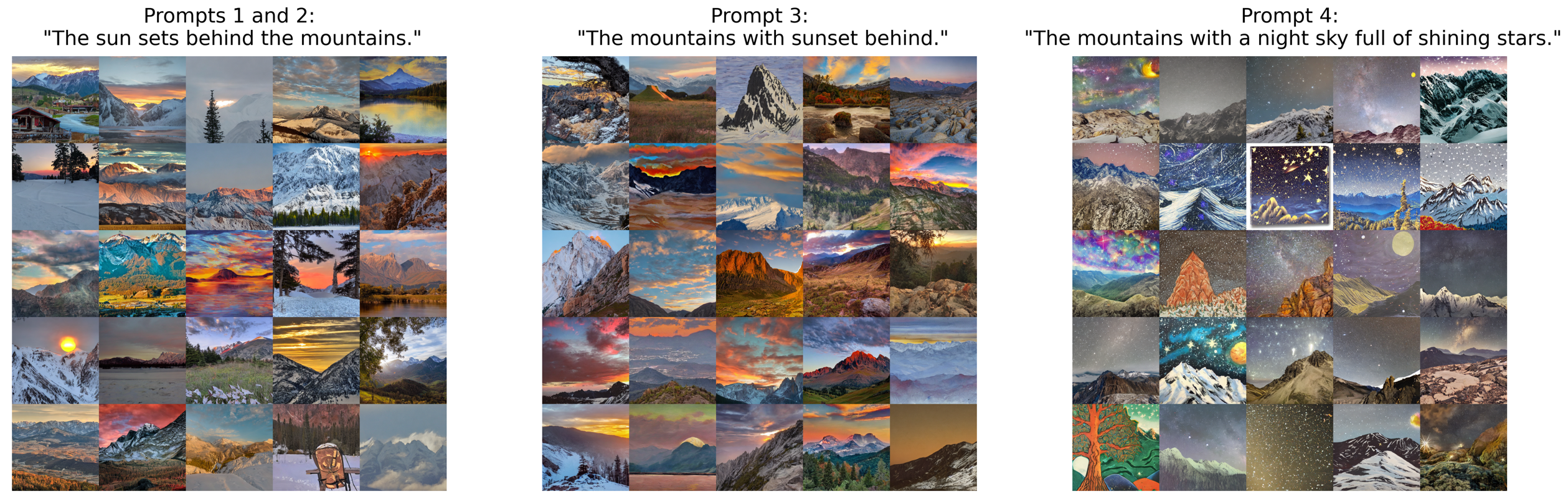}
    \caption{Generated images given different prompts by Stable Diffusion. The image size is cropped from (512, 512) to (299, 299) to accommodate the input shape for the Inception-V3 model \cite{szegedy2016rethinking}.} 
    \label{fig: SD-compare}
\end{figure}

Consider four prompts as follows: Prompt 1 - \textit{``The sun sets behind the mountains''}, Prompt 2 - \textit{``The sun sets behind the mountains''}, Prompt 3 - \textit{``The mountains with sunset behind''}, and Prompt 4 - \textit{``The mountains with a night sky full of shining stars''}. The four prompts have different levels of similarities: Prompts 1 and 2 are identical, Prompt 1 (or 2) is similar to Prompt 3, and Prompt 4 differs from all three above, with the Cosine similarity of
1, .891, .590, and .607 in Table \ref{tab:text-to-image}. Visually, images from Prompts 1 (or 2) and 3 appear very similar with only slight differences, whereas those from Prompt 4 display stars and look dramatically different, as illustrated in Fig 
\ref{fig: SD-compare}. Next, we will confirm the visual impressions through our coherence test in \eqref{hypothesis: multimodal}.

\begin{table}[!ht]
\caption{Comparison of four pairs of prompts with the cosine similarity on the
CLIP text embeddings, the FID score test statistic, and the P-value by PASS with $D=200$,
on 192-dimensional embeddings from the Inception-V3 model. }
\centering
\scalebox{.83}{
\begin{tabular}{@{}lccc@{}}
\toprule
                            & Cosine similarity &  FID score  & P-value \\ \midrule
Prompts 1 and 2 (same)      & 1.000             & ~.544                     & .990   \\
Prompts 1 and 3 (similar)   & ~.891             & 1.010                     & .124   \\
Prompts 1 and 4 (different) & ~.590             & 14.250                    & .000   \\
Prompts 3 and 4 (different) & ~.607             & 14.172                    & .000   \\ \bottomrule
\end{tabular}
\label{tab:text-to-image}}
\end{table}

To apply PAI for testing in \eqref{hypothesis: multimodal}, we construct a PASS generator using a pre-trained stable diffusion model to generate two image sets given two prompts. This pre-trained model is a well-trained state-of-the-art text-to-image model (equivalent to $n_h \to + \infty$). Then, we compute the FID score of 192-dimensional Inception-V3 embeddings between the two sets of images. To simulate the null distribution, we apply PAI to the test statistic with an MC size of $D=200$ for both image sets, where the effective size of a sample is 400.

\begin{figure}[ht]
    \centering                
    \includegraphics[scale=0.23]{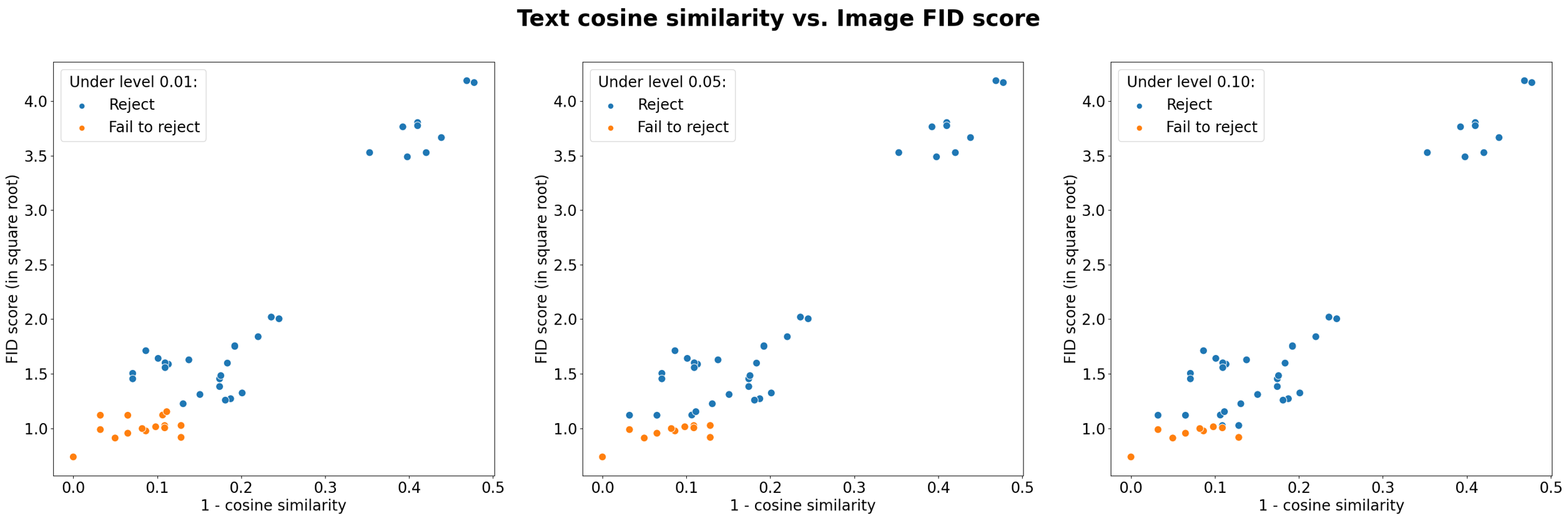}
    \caption{Pairs of FID score and Cosine similarity on embeddings generated from CLIP versus the FID score (test statistic) computed based on 192-dimensional features from the Inception-V3 model \cite{szegedy2016rethinking}, under different significance levels $\alpha = .01, .05, .1$.
    Each point in the plot represents a pair of prompts.
    } 
    \label{fig: cossim-fid}
\end{figure}

Images generated under Prompts 1 and 2, and Prompts 1 and 3, are statistically indistinguishable, given the corresponding P-values of 0.99 and 0.124 at a significance level $\alpha = .05$ in Table \ref{tab:text-to-image}. In contrast, Prompts 1 and 4, and Prompts 3 and 4 significantly differ in image generation as they have different implications. Moreover, we construct more pairs of prompts to obtain a spectrum of cosine similarity versus FID score, along with the corresponding test results. As illustrated in Fig \ref{fig: cossim-fid}, a small FID score and a large Cosine similarity imply that two prompts are conceptually equivalent or similar, which can be captured by the test under different significance levels.

\subsection{Prediction Interval}

We perform a simulation study to evaluate the accuracy and precision of prediction intervals created using PAI with a PASS generator and compare them to those obtained through the conformal method \cite{lei2018distribution}. We use a simulation model where the ground truth is accessible for assessment:
{\small
\begin{equation}\label{eq: simulation-prediction}
Y = 8 + X_1^2 + X_2 X_3 + \cos(X_4) + \exp(X_5 X_6) + 0.1 X_7 + \bm \epsilon,
\end{equation}}
where $\bm X = (X_1, \ldots, X_7)$ follows a uniform distribution over $[0,1]^7$ ($\text{Uniform}(0, 1)^7$), and $\bm \epsilon$ is normally distributed with zero mean and standard deviation $0.4 \times X_1$. We generate $3,200$ samples from \eqref{eq: simulation-prediction}, dividing them into $3,000$ for training and $200$ for testing.

To generate a conditional generative model of $Y | \bm X$, we employ a method suggested by (\cite{lugmayr2022repaint, zhang2023mixed}). Initially, we train a TabDDPM (\cite{kotelnikov2023tabddpm}) as our PASS generator on the training data to model the joint distribution of $(Y, \bm X)$. Then, we adjust the reverse process of the diffusion model for conditional generation without re-training. A predictive interval with a coverage level of $1-\alpha$ can be defined as $(l, u)$, with $l$ and $u$ being the lower $\frac{\alpha}{2}$ and upper $1-\frac{\alpha}{2}$ quantiles of the conditional distribution, estimated using the MC approach with PAI.

In our experiments, we set $\alpha = 0.05$ and compare the PAI prediction intervals against those from conformal inference. Specifically, for the latter, we split the training dataset further into a modeling sample of $2,400$ and a calibration sample of $600$. The former is used to train a CatBoost predictive model \cite{dorogush2018catboost}, while the latter helps construct conformal scores for uncertainty quantification. We evaluate the prediction intervals of both methods on the test sample.

\begin{figure}[ht]
    \centering
    \includegraphics[width=\linewidth]{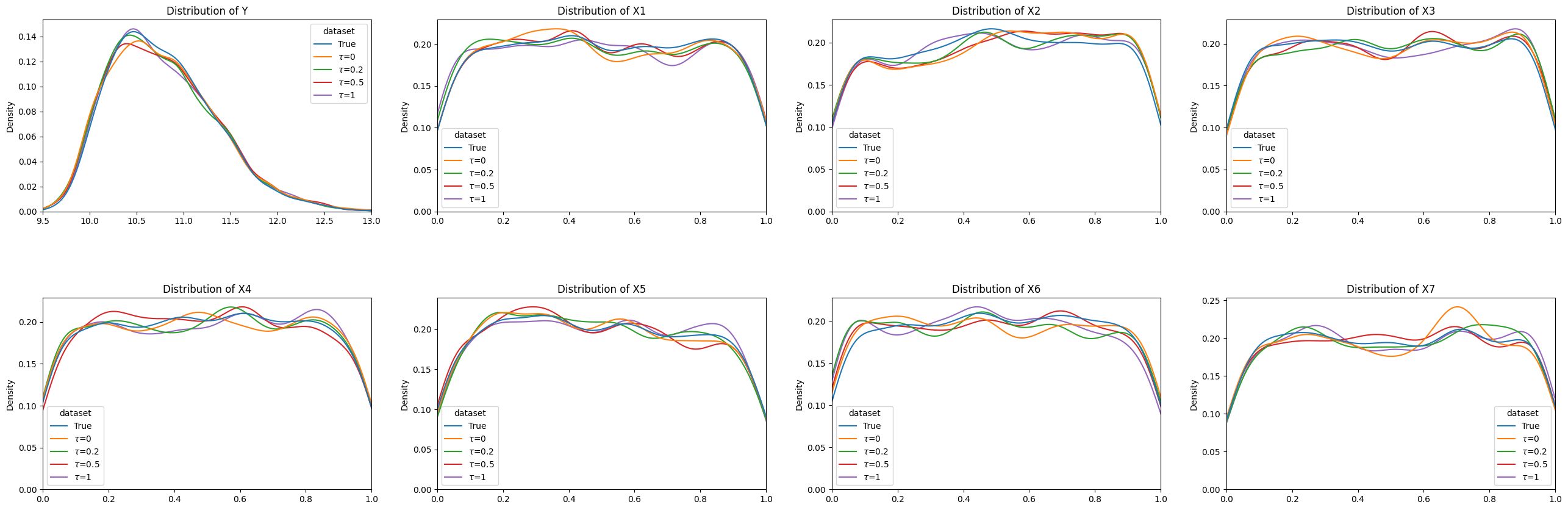}
\caption{Kernel density estimates (KDE) of marginal distributions for $(Y, \bm X)$ as learned by PASS for perturbation sizes $\tau \in {0, 0.2, 0.5, 1}$, compared with those from an independent evaluation sample of size $3,000$ from the underlying true distribution (blue).}
    \label{fig: simulation-dist}
\end{figure}

Here, we highlight that the sizes of perturbations do not compromise the validity or accuracy of the learned distribution, due to using distribution-preserving perturbation functions, c.f.,   \eqref{DPG}. This claim is reinforced by the results depicted in Figure \ref{fig: simulation-dist}, demonstrating that the distribution learned by the PASS algorithm remains consistent across various perturbation sizes $\tau \in \{0, 0.2, 0.5, 1\}$, closely matching the true underlying distribution. Additional validation comes from the data presented in Table \ref{tab:dist-distance}, which shows negligible variation in distributional distances
under the 1- and 2-Wasserstein distances \footnote{\url{https://pythonot.github.io/quickstart.html\#computing-wasserstein-distance}}, and Fr\'echet Inception Distance (FID)\footnote{Note that FID is 2-Wasserstein distance under Gaussian assumption.} for different perturbation sizes, all suggesting comparable generative error rates. In conclusion, the perturbation size only does not affect PAI, which utilizes the MC simulation method.

\begin{figure}[ht]
    \centering
    \includegraphics[width=\linewidth]{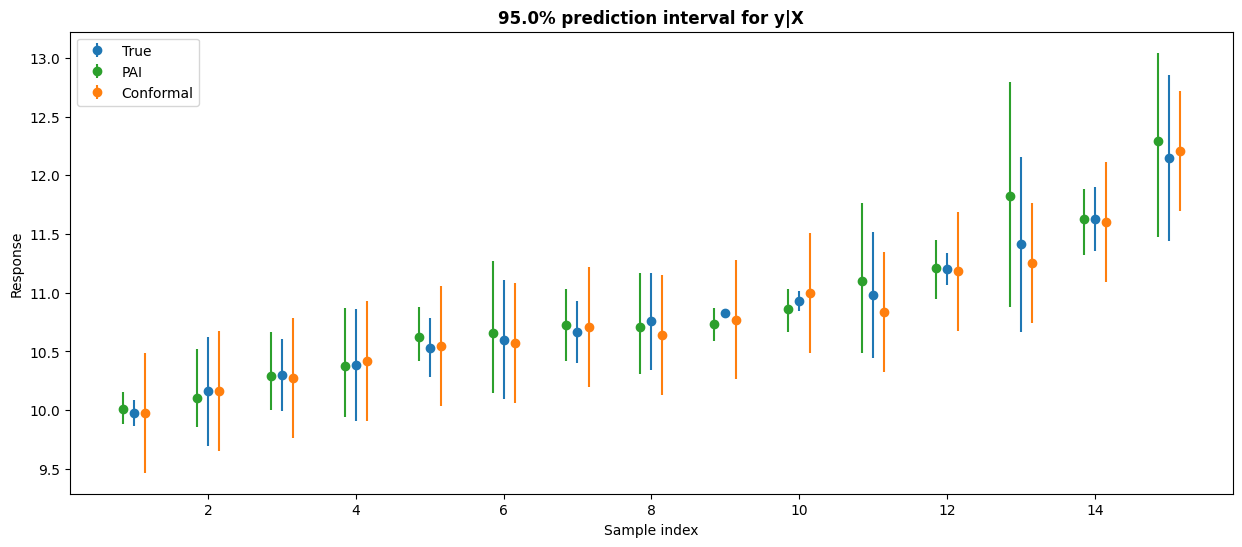}
\caption{Comparison of the $95\%$ prediction intervals obtained using PAI (depicted in green), conformal inference (depicted in orange), and the actual observed values (depicted in blue), for a randomly selected subset of $15$ data points from the test set.}
    \label{fig: PIs}
\end{figure}

About $68\%$ of the intervals using PAI are found to be shorter than those obtained via conformal inference, as depicted in Figure \ref{fig: PIs}, where PAI intervals are contrasted with those from conformal inference and the actual values on randomly selected test points. PAI intervals also show a better alignment with the true values, highlighting PAI's effectiveness as a non-parametric inference method.

\begin{figure}[ht]
    \centering
    \includegraphics[width=\linewidth]{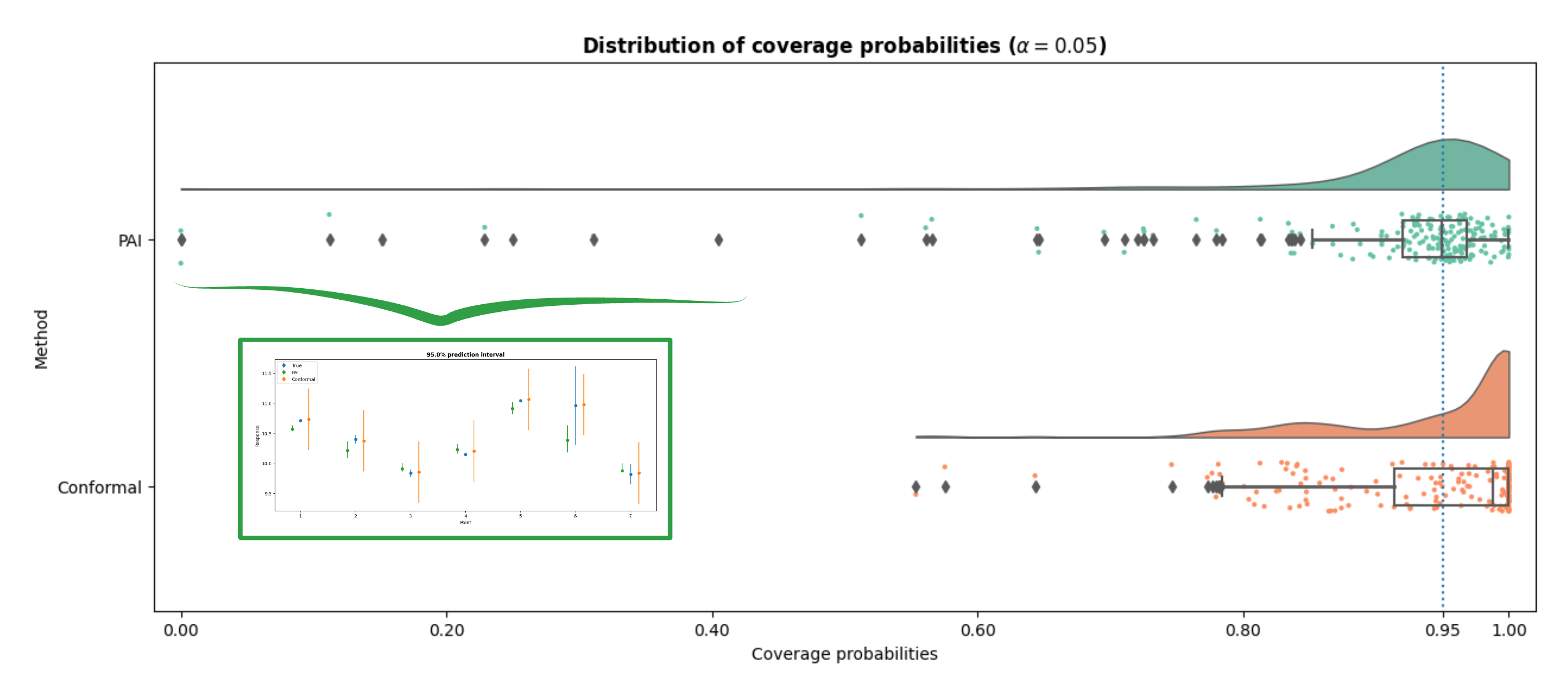}
\caption{Comparison of the coverage probability distributions obtained using PAI (in green) and conformal inference (in orange), based on $200$ test points. The plot is divided into two sections for each method: the upper section displays the Kernel Density Estimate (KDE) of the probabilities, while the lower section presents the boxplot of the distribution. Additionally, an inset within the plot highlights the prediction intervals for outliers identified by PAI.}
    \label{fig: coverage-probs}
\end{figure}

Furthermore, PAI prediction intervals maintain accurate coverage probabilities. As illustrated in Figure \ref{fig: coverage-probs}, while conformal inference intervals tend to be wider and more conservative, PAI intervals achieve nearly exact coverage: their median coverage probability is $0.95$, consistent with the specified level. However, PAI's average coverage probability is slightly lower at $0.9$ due to outliers in the underlying model with small variance and some bias in the PASS generator, which slightly mis-aligns the prediction intervals' centers, despite the estimated lengths being close to the actual values.

\begin{table}[ht]
\centering
\caption{Distributional distances between the synthetic sample and an evaluation sample from the underlying true distribution, each of size $3,000$. Parenthetical numbers represent the standard errors derived from repeated experiments.}
\resizebox{\columnwidth}{!}{%
\begin{tabular}{@{}clllc@{}}
\toprule
              & $\tau = 0.0$  & $\tau = 0.2$  & $\tau = 0.5$  & $\tau = 1.0$  \\ \midrule
FID           & 0.024 (0.005) & 0.023 (0.005) & 0.023 (0.005) & 0.024 (0.005) \\
1-Wasserstein & 1.238 (0.004) & 1.237 (0.004) & 1.238 (0.005) & 1.238 (0.005) \\
2-Wasserstein & 1.298 (0.005) & 1.296 (0.004) & 1.289 (0.005) & 1.298 (0.006) \\ \bottomrule
\end{tabular}%
}
\label{tab:dist-distance}
\end{table}

\section{Conclusion}
This paper introduces PAI, a novel inference framework grounded in a generative scheme, PASS, which facilitates statistical inference from complex and unstructured data types such as images and texts. PAI addresses the lack of effective uncertainty quantification methods in black-box models like deep neural networks.

The PAI framework, building on PASS, specializes in estimating the distribution of statistics through Monte Carlo experiments, offering a robust method for statistical inference. A key strength of PAI is its theoretical guarantee of inference validity, even in scenarios of scarce data. This paper demonstrates its broad applicability.

Nonetheless, PAI has its limitations. Its primary challenge is the computational demand during Monte Carlo experiments. Also, PAI's performance and accuracy largely depend on the effectiveness of PASS.

On the other hand, PASS utilizes generative models, such as diffusion models and normaliing flows, to mirror the raw data distribution. It can also harness large pre-trained generative models to enhance estimation accuracy. PASS's generator supports data integration and personalization through multivariate rank matching on latent variables, maintaining privacy via data perturbation. Theoretically, we explore PASS's sampling properties, confirming the approximation of latent variable ranks post-data perturbation. Experimental results highlight PASS's superior generation quality.

Our primary goal is to provide researchers with tools that foster reliable and reproducible conclusions from data. These tools have the potential to enhance the credibility and reliability of data-driven discoveries and statistical inferences.

\bibliographystyle{ieeetr}
\bibliography{preref-1.bib,diffpriv-1.bib}

\newpage

\begin{IEEEbiography}[{\includegraphics[width=1in,height=1.25in,clip,keepaspectratio]{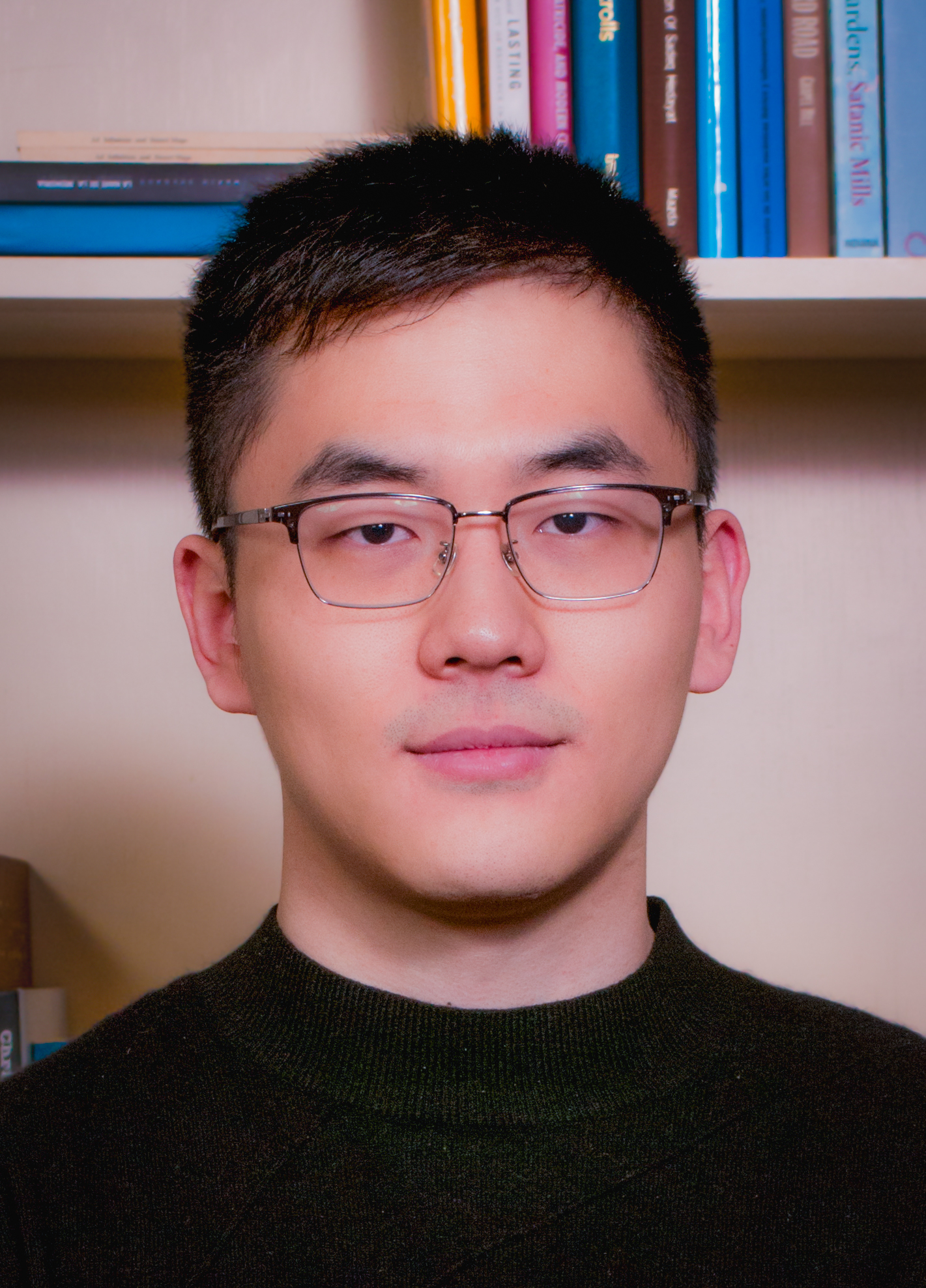}}]{Yifei Liu} received a B.S. degree in Statistics
from University of Science and Technology of China, China, in 2019.  He is currently a Ph.D. candidate in Statistics at the University of Minnesota, Minneapolis, USA.
His research interests include statistical inference for black-box models, generative models including diffusion models and normalizing flows, statistical learning methods and theory, and deep learning applications.
\end{IEEEbiography}

\begin{IEEEbiography}[{\includegraphics[width=1in,height=1.25in,clip,keepaspectratio]{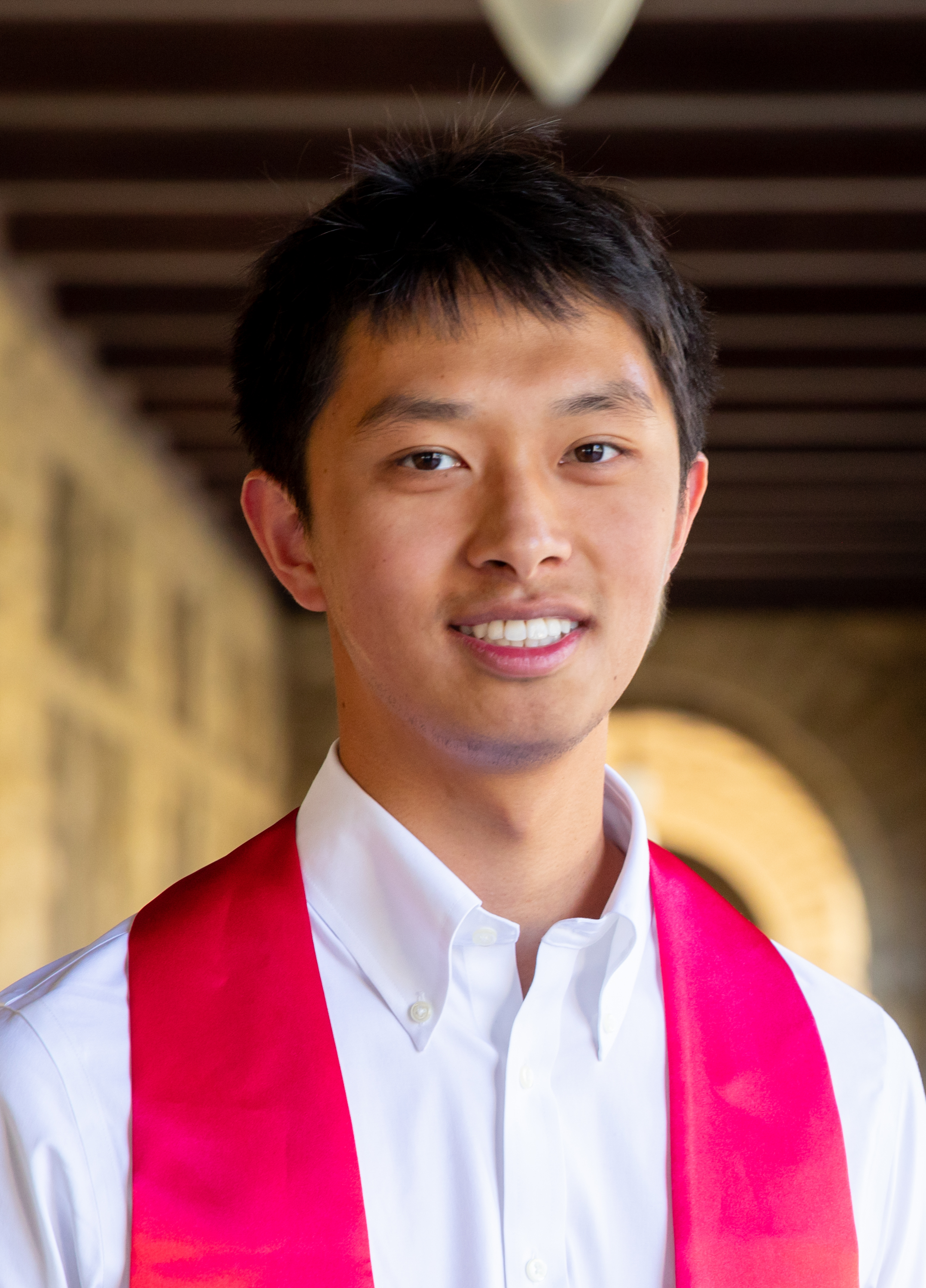}}]{Rex Shen} received a B.S. degree in Mathematics and Computational Science and a M.S. degree in Statistics from Stanford University, USA, in 2022. He is currently a Ph.D. candidate in Statistics at Stanford University.  His current research interests span synthetic data generation, Generative AI and its applications. 
\end{IEEEbiography}

\begin{IEEEbiography}[{\includegraphics[width=1in,height=1.25in,clip,keepaspectratio]{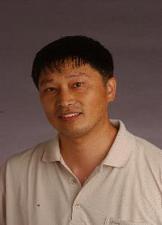}}]{Xiaotong Shen} received a B.S. degree in Applied Mathematics from Peking University, Beijing, China, in 1985, and a Ph.D. degree in Statistics from the University of Chicago, Chicago, USA, in 1991. Currently, He holds the position of the John Black Johnston Distinguished Professor in the School of Statistics at the University of Minnesota, Minneapolis, USA. His research interests include machine learning and data science, high-dimensional inference, non/semi-parametric inference, causal relations, graphical models, explainable Machine Intelligence, personalization, recommender systems, natural language processing, generative modeling, and nonconvex minimization. He is a fellow of the American Statistical Association, the American Association for the Advancement of Science, and the Institute of Mathematical Statistics.
\end{IEEEbiography}

\hyphenation{op-tical net-works semi-conduc-tor IEEE-Xplore}
\def\BibTeX{{\rm B\kern-.05em{\sc i\kern-.025em b}\kern-.08em
    T\kern-.1667em\lower.7ex\hbox{E}\kern-.125emX}}


\def\spacingset#1{\renewcommand{\baselinestretch}%
{#1}\small\normalsize} \spacingset{1}


\newcommand {\pv} {p_{\scalebox{1}{$\scriptscriptstyle \bm{V}$}}}
\newcommand {\hpv} {\tilde p_{\scalebox{1}{$\scriptscriptstyle \bm{V}$}}}




\title{Supplementary Materials for Novel Uncertainty Quantification through Perturbation-Assisted Sample Synthesis}

\author{
Yifei Liu, Rex Shen, Xiaotong Shen\textsuperscript{\orcidlink{0000-0003-1300-1451}}
  \thanks{Yifei Liu is with the School of Statistics, University of Minnesota, MN, 55455 USA 
  (email: liu00980@umn.edu).}
\thanks{Rex Shen is with the Department of Statistics, Stanford University, CA, 94305 USA (email:rshen0@stanford.edu ).}
\thanks{Xiaotong is with the School of Statistics, University of Minnesota, MN, 55455 USA (email: xshen@umn.edu).}
}

\maketitle



\section{Overview}
This document is structured as follows: Section \ref{training details} contains implementation details for the examples in Section \ref{numerical} of the paper.
Section \ref{appendix: multivariate rank} discusses the application of multivariate ranks to practical rank matching in PASS. 
Section \ref{sec: rank preserve} details the theoretical basis for distribution-invariant perturbation that maintains approximate rank preservation, with proofs in Section \ref{appendix: proof of rank preservation}. Section \ref{appendix: proof of main results} is devoted to
proofs related to the sampling properties of PASS, PAI validity, and pivotal inference. Finally, Section \ref{appendix: learning theory} covers the learning theory for normalization flows.

\section{Implementation Details} \label{training details}

The codes used for the numerical experiments presented in this paper are available in the following repository: \href{https://github.com/yifei-liu-stat/pass}{https://github.com/yifei-liu-stat/pass}.

\subsection{Image Synthesis}\label{training: eg1}

\begin{algorithm}[ht]\label{algorithm: eg1}
    \setcounter{AlgoLine}{0}
    \LinesNumbered
    \SetKwInOut{Input}{Input}
    \SetKwInOut{Output}{Output}
    \SetKwInOut{Result}{Result}
    \SetKwInOut{Initialize}{Initialize}
    \caption{PAI: Image Synthesis Inference}
    
    \Input{Test image embeddings $\mathbb{S}_{\text{test}}$; Generated embeddings $\mathbb{S}_{\text{fake}}$ from a candidate generator $\mc G$; PASS generator $\mc G_h$; MC size $D$. }
    \Output{A P-value indicating whether $\mc G$ has similar performance compared to $\mc G_h$ by testing \eqref{hypothesis1}.}

    \BlankLine
    \tcp{Calculate the test statistic}
    Compute $T = \text{FID}(\hat P_{\text{test}}, \hat P_{\text{fake}})$ where $\hat P_{\text{test}}$ and $\hat P_{\text{fake}}$ are empirical distributions of $\mathbb S_{\text{test}}$ and $\mathbb{S}_{\text{fake}}$.\\
    \tcp{Estimate the distribution of $T$ under $H_0$}
    \For{$d = 1, \dots, D$}{
        Generate $\mathbb S^{(d)}$ of size $|\mathbb S_{\text{fake}}|$ using $\mc G_h$;\\
        Compute $T^{(d)} = \text{FID}(\hat P_{\text{test}}, \hat P^{(d)})$, where $\hat P^{(d)}$ is the empirical distributions of $\mathbb S^{(d)}$.
    }
    Compute $F_D$, the empirical CDF of $(T^{(d)})_{d = 1}^D$.\\
    \tcp{Empirical two-sided P-value}
    \textbf{return} P-value = $2 \min(F_D(T), 1 - F_D(T))$.
\end{algorithm}

We undertake training of various generative models to accurately learn the CIFAR-10 dataset's distribution, leveraging either a single TITAN RTX GPU or a configuration involving multiple GPUs. To support our numerical experiments, we implemented modifications in the source code. It should be noted that the optimization of tuning parameters for improving the quality of generation was not our focus. Instead, our main objective revolves around using these trained models for uncertainty quantification.

\begin{itemize}

\item PASS and DDPM: To train diffusion models,
we adopt the code \texttt{Denoising-Diffusion}\footnote{\href{https://github.com/lucidrains/denoising-diffusion-pytorch}{DDPM: https://github.com/lucidrains/denoising-diffusion-pytorch}} using the Adam optimizer with a learning rate of $8 \times 10^{-5}$, $4\times 10^5$ time steps, a batch size of 32, and 250 sampling time steps using a light-weight fast sampler DDIM \cite{song2020denoising}.

\item DCGAN: We adopt the implementation of \texttt{DCGAN-CIFAR10}\footnote{\href{https://github.com/Amaranth819/DCGAN-CIFAR10-Pytorch}{DCGAN: https://github.com/Amaranth819/DCGAN-CIFAR10-Pytorch}} using the Adam optimizer with a learning rate of $1\times 10^{-3}$, a batch size of 128, and 30 epochs. Both the generator and discriminator are CNNs and are trained alternatively.

\item GLOW: We implement using \texttt{Glow-PyTorch}\footnote{\href{https://github.com/y0ast/Glow-PyTorch}{Glow-PyTorch: https://github.com/y0ast/Glow-PyTorch}}. Specifically, we train the GLOW model with its default normalizing flow structure using an additive coupling layer and the Adamax optimizer with a learning rate of $5 \times 10^{-4}$, weight decay of $5 \times 10^{-5}$, a batch size of 64, and 250 epochs.

\end{itemize}

\subsection{Sentiment Word Inference}\label{training: eg2}

\begin{algorithm}[ht]\label{algorithm: eg2}
    \setcounter{AlgoLine}{0}
    \LinesNumbered
    \SetKwInOut{Input}{Input}
    \SetKwInOut{Output}{Output}
    \SetKwInOut{Result}{Result}
    \SetKwInOut{Initialize}{Initialize}
    \caption{PAI: Sentiment Word Inference}
    
\Input{Words of interest $\mc W_M$; Pre-trained transformer encoder $\mc E$; Attention threshold $\delta \in (0, 1)$; Inference sample $\mathbb{S}$; PASS generator $\mc G_h$ on embeddings; MC size $D$. }
    \Output{A P-value indicating whether $\mc W_M$ contributes to classification by testing \eqref{hypothesis2}.}

    \BlankLine
    \tcp{Create an inference sample masked by $\mc W_M$}
    \For{every sequence $\mb z = (z_t)_{t = 1}^T$ in $\mathbb S$}{
        \For{every token $z_t$ in $\mb z$}{
            Use $\mc E$ to extract attention weights of all other tokens in $\mb z$, attended by $z_t$;\\
            Mask $\delta$ proportion of all other tokens starting from those with small attention weights;\\
            Collect the masked sequence $\mb z_M$.
        }
    }
    Get the masked inference sample $\mathbb 
    S_M$ from $\mb z_M$.
    Then apply $\mc E$ to inference samples $\mathbb S$ and $\mathbb S_M$, and get embedded inference samples $\mathbb E$ and $\mathbb E_M$.\\
    \tcp{Calculate test statistic with embeddings}
    Compute $T$ using \eqref{tests} on $\mathbb E$ and $\mathbb E_M$.\\
    \tcp{Estimate the distribution of $T$ under $H_0$}
    \For{$d = 1, \dots, D$}{
        Generate $\mathbb E^{(d)}$ of size $|\mathbb E|$  using $\mc G_h$;\\
        Compute $T^{(d)}$ on $\mathbb E^{(d)}$ using \eqref{tests}.
    }
    Obtain $F_D$, the empirical CDF of $(T^{(d)})_{d = 1}^D$.\\
    \textbf{return} P-value = $F_D(T)$.
\end{algorithm}

We conduct the training of a sentiment classifier and implement normalizing flows as follows.

\begin{itemize}

\item{Sentiment Classifier.} We fine-tuned a pre-trained DistilBERT model from the \texttt{transformers} library\footnote{\href{https://github.com/huggingface/transformers}{transformers: https://github.com/huggingface/transformers}} by attaching a classification head, where we fine-tuned for five epochs with an Adam optimizer using a learning rate of $10^{-5}$ and default exponential decay rates.

\item{Normalizing Flows.} We trained normalizing flows to learn the null distribution using a RealNVP architecture with ten affine coupling layers, with batch normalization and reverse permutation layers in between. We learned each affine coupling layer's parameters using a ReLU neural network with three hidden layers with dimensions 8192, 4096, and 2048. The model was implemented using the \texttt{pyro} framework\footnote{\href{https://github.com/pyro-ppl/pyro}{pyro: https://github.com/pyro-ppl/pyro}}. The normalizing flow model was trained using an Adam optimizer with a learning rate of $7 \times 10^{-5}$ and default exponential decay rates. The batch size was set to 64, and we used a linear one-cycle learning rate scheduler \cite{smith2019super} for 500 epochs to avoid overfitting while allowing for warm starting. The training process also involved monitoring the generated embeddings versus true embeddings for selected coordinate pairs.

\end{itemize}

All training processes, including DistilBERT's fine-tuning full training of normalizing flows, are performed on 4 TITAN RTX GPUs with mixed precision.

\subsection{Text-to-Image Generation} \label{training: eg3}

\begin{algorithm}[ht]\label{algorithm: eg3}
    \setcounter{AlgoLine}{0}
    \LinesNumbered
    \SetKwInOut{Input}{Input}
    \SetKwInOut{Output}{Output}
    \SetKwInOut{Result}{Result}
    \SetKwInOut{Initialize}{Initialize}
    \caption{PAI: Multimodal Inference}
    
    \Input{Two sets of image embeddings $\mathbb S_1$ and $\mathbb S_2$ generated from two prompts $\mb x^{(1)}$ and $\mb x^{(2)}$; PASS generator $\mc G_h$; MC size $D$. }
    \Output{A P-value indicating whether $\mb x^{(1)}$ and $\mb x^{(2)}$ generate similar images by testing \eqref{hypothesis: multimodal}.}

    \BlankLine
    \tcp{Calculate the test statistic}
    Compute $T = \text{FID}(\hat P_1, \hat P_2)$ where $\hat P_1$ and $\hat P_2$ are empirical distributions of $\mathbb S_1$ and $\mathbb S_2$.\\
    \tcp{Estimate the distribution of $T$ under $H_0$}
    \For{$k = 1, 2$}{
        \For{$d = 1, \dots, D$}{
            Generate $\mathbb S^{(d)}_1$ of size $|\mathbb S_1|$ and $\mathbb S^{(d)}_2$ of size $|\mathbb S_2|$ using $\mc G_h$ given prompt $\mb x^{(k)}$;\\
            Compute $T^{(d)}_k = \text{FID}(\hat P_1^{(d)}, \hat P_2^{(d)})$, where $\hat P_1^{(d)}$ and $\hat P_2^{(d)}$ are the empirical distributions of $\mathbb S^{(d)}_1$ and $\mathbb S^{(d)}_2$, respectively.
        }
    }
    Get $F_D$, the empirical CDF of $(T^{(d)}_1)_{d = 1}^D \cup (T^{(d)}_2)_{d = 1}^D$.\\
    \textbf{return} P-value = $1 - F_D(T)$.
\end{algorithm}

For a text-to-image generator, we use the pre-trained stable diffusion model v1-4 hosted on HuggingFace\footnote{\href{https://huggingface.co/CompVis/stable-diffusion}{stable-diffusion:  https://huggingface.co/CompVis/stable-diffusion}}, with 50 inference steps and guidance scale as 2. To calculate the FID score, we utilize the implementation from \texttt{torchmetrics}\footnote{\href{https://github.com/Lightning-AI/torchmetrics}{torchmetrics: https://github.com/Lightning-AI/torchmetrics}}, which uses pre-trained Inception-V3 model \cite{heusel2017gans} for extracting image embeddings. The experiment is performed on 8 TITAN RTX GPUs with mixed precision.

\subsection{Prediction Interval}

For the diffusion model applied to tabular data, our approach aligns with the framework outlined in TabDDPM 
\cite{kotelnikov2023tabddpm}, employing the quantile transformation to map each numerical column into a standard 
Gaussian distribution. The diffusion process is governed by a linear schedule, with the parameters of noise 
scheduler $\beta_{min}=10^{-4}$ and $\beta_{max}=0.02$ across $1,000$ diffusion steps. The architecture centers on a multilayer perceptron with time embedding for the noise prediction network.

Optimization leverages the Adam optimizer, set to a learning rate of $10^{-3}$, and the model trains over $1,000$ epochs, incorporating early stopping to enhance efficiency. All training and sampling activities utilize a single TITAN RTX GPU.

For conformal inference, a predictive model is developed with CatBoost \footnote{\href{https://github.com/catboost/catboost}{CatBoost: https://github.com/catboost/catboost}} over $1,000$ iterations, also applying early stopping. The normalized deviation serves as the conformal score to establish prediction intervals, with further methodological details available in \cite{angelopoulos2021gentle}.

\section{Multivariate Rank} \label{appendix: multivariate rank}

We now discuss the optimal transportation approach to define ranks for a random vector, referred to as \textit{multivariate ranks} \cite{deb2023multivariate}, which generalizes the concept of univariate rank. First, we introduce some notations: for two probability measures $\mu$ and $\nu$ on $\mathcal{R}^d$, and a function $F: \mathcal{R}^d \to \mathcal{R}^d$, if for any $\X \sim \mu$, $F(\X) \sim \nu$, we say $F$ pushes $\mu$ to $\nu$, denoted as $F\#\mu = \nu$. Here we use $P(\mathcal{R}^d)$ 
and $\mathcal{U}^d$ to denote the set of all probability measures defined on $\mathcal{R}^d$ and
for the uniform measure on $[0, 1]^d$.

\begin{definition}[Population Rank Map]\label{def: population-rank-map} For any $\mu \in P(\mathcal{R}^d)$, the map $R: \mathcal{R}^d \to [0, 1]^d$ is the \textit{rank map} of $\mu$ if (i) $R\#\mu = \mathcal{U}^d$ and (ii) $R = \nabla \Psi$ for some real-valued convex function $\Psi$ on $\mathcal{R}^d$. 
\end{definition}

\begin{remark} Requirement (ii) in Definition \ref{def: population-rank-map} ensures, but is not limited to, the monotonicity property of the map. For example, in the univariate case, if $x < y$, then it is natural to expect their ranks to satisfy $R(x) < R(y)$. In the multivariate case, we do not have a canonical ordering, but (ii) implies that $R(\cdot)$ is cyclically monotone \cite{rockafellar1997convex}, which is a generalization of univariate monotonicity. 
\end{remark}

The rank map $R(\cdot)$ is well-defined by McCann's Theorem (\cite{deb2023multivariate,mccann1995existence,villani2021topics}). 
Moreover, if $\mu$ has a finite second moment, then 
$R(\cdot)$ is also the solution of \textit{Monge's problem} $\inf_{F: F\#P = Q} \int c(\mathbf{x}, \mathbf{F}(\mathbf{x})) dP$ with $P = \mu$, $Q = \mathcal{U}^d$, and $c(\mathbf{y}, \mathbf{z}) = \|\mathbf{y} - \mathbf{z}\|^2$. Note that the rank map reduces to the CDF of the measure $\mu$ when $d = 1$ but is generally not analytically tractable when $d > 1$ except for some special cases. In practice, we estimate the population rank map via a random sample.

\begin{definition}[Empirical Rank Map] Denote $\mb h_1^d, \ldots, \mb h_n^d$ as a $d$-dimensional Halton sequence with $\nu_n$ as its empirical measure. Let $\mu_n$ be the empirical measure based on $\Y_1, \ldots, \Y_n$.
Then, the map $R_n^{\Y}: (\Y_i)_{i = 1} \to (\mb h_i^d)_{i = 1}^n$ obtained by solving Monge's problem with $P = \mu_n$ and $Q = \nu_n$, is called the \textit{empirical rank map} of $(\Y_i)_{i = 1}^n$. \end{definition}

\begin{remark} The Halton sequence \cite{berblinger1991monte} is a fixed sequence in $[0, 1]^d$ whose empirical measure converges weakly to the uniform distribution $U[0,1]^d$. This sequence is the multivariate ranks in $\mathcal{R}^d$. \end{remark}

 The computation of empirical ranks $R_n^{\Y}$ amounts to solving the discrete Monge's problem, 
or identifying a permutation map $\pi_* = \argmin_{\pi} \sum_{i = 1}^n \|\Y_{\pi(i)} - \mb h_i^d\|_2^2$ in all permutation maps $\pi$ on $\{1, \ldots, n\}$. Then, $R_n^{\Y}(\Y_{\pi_*(i)}) = \mb h_i^d$.
In the case of $d = 1$, we choose $h_i^1 = \frac{i}{n}$; $i = 1, \ldots, n$, and the empirical rank map $R_n^{\Y}$ coincides with the univariate rank up to a constant.

\begin{algorithm}[t]\label{algorithm: rank matching}
    \setcounter{AlgoLine}{0}
    \LinesNumbered
    \SetKwInOut{Input}{Input}
    \SetKwInOut{Output}{Output}
    \SetKwInOut{Result}{Result}
    \SetKwInOut{Initialize}{Initialize}
    \caption{Multivariate rank matching}
    
    \Input{Latent representations $(T(\mb Z_i))_{i = 1}^n$; $(\mb U_i)_{i = 1}^n$.}
    \Output{A permutation map $r$ that matches the ranks of $(\mb U_{r(i)})_{i = 1}^n$ and those of $(T(\mb Z_i))_{i = 1}^n$.}

    \BlankLine
    Compute the permutation map $\pi_1(\cdot)$ by minimizing $\sum_{i = 1}^n 
\|T(\Z_{\pi(i)}) - \bm h_i^d\|_2^2$ in all permutation maps;\\
    Compute the permutation map $\pi_2(\cdot)$ by minimizing $\sum_{i = 1}^n \|\bm U_{\pi(i)} - \bm h_i^d\|_2^2$ in all permutation maps;\\
    \textbf{return} $r = \pi_2 \circ \pi_1^{-1}$.
\end{algorithm}

{\bf Rank matching in PASS.}{
The empirical ranks  
provides a distribution-free way to measure the closeness between two multivariate samples in 
terms of their ranks. This property motivates the approximate rank matching in \eqref{DPG} for PASS.
To determine a permutation map $r=\pi_2 \circ \pi_1^{-1}$ matching the empirical ranks of $(\U_{r(i)]})_{i = 1}^n$ to those of $(T(\Z_i))_{i = 1}^n$,
we compute the permutation map $\pi_1(\cdot)$ by minimizing $\sum_{i = 1}^n 
\|T(\Z_{\pi(i)}) - \bm h_i^d\|_2^2$ and $\pi_2(\cdot)$ by minimizing $\sum_{i = 1}^n \|\bm U_{\pi(i)} - \bm h_i^d\|_2^2$. 
}

{\bf Computation complexity.} The computational complexity for $r(\cdot)$ boils down to solving a balanced linear sum assignment problem (LSAP) with cost matrix $\mathbf{C}=(C_{ij})_{n \times n}$, where $C_{ij} = n^{-1} \|\Y_i - \mb h_j^d\|_2^2$. This problem can be solved using the Hungarian algorithm \cite{jonker1987shortest}. When $n \gg d$, the complexity of 
matching multivariate ranks in \eqref{DPG} is $O(n^3)$ as
many variants of the Hungarian algorithm admit a worst-case complexity of $O(n^3)$. 

\section{Rank-Preserving Perturbation Function}\label{sec: rank preserve}

In \eqref{DPG}, we consider a specific type of perturbation function $W$ to approximately preserve 
the multivariate ranks of $(T(\mb Z_i))_{i = 1}^n$, that is, $(\mb V_i)_{i = 1}^n$ and $(T(\mb Z_i))_{i = 1}^n$ share 
similar multivariate ranks.
Specifically, $\V_i = W(\mb U_{r(i)} + \mb e_i)$ in \eqref{DPG}, where $W$ is the optimal transport map from $F_{\tilde \U} \equiv F_{\U} \otimes F_{\e}$ to $F_{\U}$, which is the solution to Monge's problem, and $\mb e_i = \tau \mb \epsilon_i$, and $\tau > 0$ is the perturbation size and $\mb \epsilon_i \sim F_{\mb \epsilon}$ is a standardized noise distribution.
Note that by \eqref{DPG}, $\V, \U, T(\Z) \sim F_{U}$, and thus we do not distinguish between $F_{\V}$, $F_{T(\Z)}$ and $F_{\U}$.
As indicated by
Theorem \ref{theorem: rank preservation}, $W$ achieves the desired goal of rank preservation.

To proceed, denote $f(\tau) = \sup_{\bm z \in \mc Z} \|R(\bm z) - \tilde R(\bm z)\|^2$ as the discrepancy between base distribution $F_{\U}$ and linearly perturbed distribution $F_{\tilde \U}$, where $R$ and $\tilde R$ are the corresponding population rank maps.
Let $\mb R^{\V} = [\mb R^{\V}_1, \ldots, \mb R^{\V}_n]$ and $\mb R^{\mb T} = [\mb R^{\mb T}_1, \ldots, \mb R^{\mb T}_n]$ be the multivariate ranks of $(\V_i)_{i = 1}^n$ and those of 
$(T(\Z_i))_{i = 1}^n$, respectively.
$\mathbb J_g$ stands for the Jacobian of a function $g$.
Define $\|\mb R^{\V} - \mb R^{\mb T}\|_n = \Big(n^{-1} \sum_{i = 1}^n \|\mb R^{\V}_i - \mb R^{\mb T}_i\|_2^2\Big)^{1/2}$.
Let the perturbed base vector $\tilde \U = \U + \e \sim F_{\tilde \U}$. We assume the following conditions:

\begin{itemize}
    \item[(A)] (Distribution) $F_{\U}$ and $F_{\tilde \U}$ are absolutely continuous on $\mc R^d$.
    \item[(B)] (Moment) For some $t > 0$, $\alpha > 0$ and $\tau_0 > 0$, $\E[\exp(t \|\U\|^\alpha)] < \infty$ and $ \sup_{\tau \in [0, \tau_0]} E[\exp(t \|\tilde \U\|^\alpha)] < \infty$.
    \item[(C)] (Lipschitz) Population rank maps $R$ and $\tilde R$ are $L$-Lipschitz continuous.
    \item[(D)] (Commutability) {\small $J_W(\tilde \U) J_R(W(\tilde \U)) = J_R(W(\tilde \U)) 
J_W(\tilde \U)$} almost surely.
\end{itemize}

\begin{theorem}[Rank-preservation perturbation]\label{theorem: rank preservation}
Under Assumptions (A)-(D), for $d \geq 1$, 
{\small 
\begin{equation*}
    \|\mb R^{\V} - \mb R^{\mb T}\|_n^2 = O_p\paren{\max\paren{r_{n, d} \paren{\log n}^{t_{\alpha, d}}, \tau^2 n^{-1/2} + f(\tau)} }, 
\label{eq: rank-preservation-rate}
\end{equation*}
}
where $r_{n, d} = n^{-1}$ and $t_{\alpha, d}=0$ when $d = 1$; $r_{n, d} = n^{-1/2}$ and $t_{\alpha, d}=
(4\alpha)^{-1}\paren{4 + 0 \vee (2\alpha + 2d\alpha - d)}$ when $d = 2, 3$ and 
$t_{\alpha, d}=\alpha^{-1} \vee (7/2)$ when $d = 4$; $r_{n, d} = n^{-1/d}$ and $t_{\alpha, d}=2(1 + d^{-1})$ when $d \geq 5$.
\end{theorem}

\noindent{\bf Proof of Theorem \ref{theorem: rank preservation}:} 
Note that
\begin{equation}\label{eq: decomposition}
    \|\bm R^{\V} - \bm R^{\mb T}\|^2_n \leq 2 \|\bm R^{\V} - \tilde{\mb R}\|^2_n + 2  \|\tilde{\mb R} - \bm R^{\mb T}\|^2_n,
\end{equation}
where $\tilde{\mb R} = [\tilde{\mb R}_1, \dots, \tilde{\mb R}_n]$ are the empirical ranks of perturbed samples $(\tilde \U_i)_{i = 1}^n$ with $\tilde \U_i = \bm U_{r(i)} + \mb e_i$.
Hence, the desired result follows by applying Lemma \ref{lemma: rank preservation of optimal transport} to the first term on the RHS of \eqref{eq: decomposition}, and Lemma \ref{lemma: stability of empirical rank maps} to the second one by noticing that $\mb R^{\mb T}$ are also empirical ranks of $(\U_{r(i)})_{i = 1}^n$. \hfill $\square$

Theorem \ref{theorem: rank preservation} suggests that $(\mb V_i)_{i = 1}^n$ preserves 
the multivariate ranks of $(T(\mb Z_i))_{i = 1}^n$ when $n \to \infty$, $\tau \to 0^+$  and $f(\tau) \to 0$ as $\tau \to 0^+$.
An illustrative example for $W$ that satisfies Assumptions (A)-(D) is provided next.

\begin{example} \label{ex1} Let $F_{\U} = \otimes_{j=1}^d F_j$ and $F_{\tilde \U} = \otimes_{j=1}^d \tilde F_j$. 
Assume that $F_j$ and $\tilde F_j$ have marginal densities $f_j$ and $\tilde f_j$ satisfying $\max_{j=1, \dots, d}(\|f_j \vee \tilde f_j\|_{\infty}) \leq L$.
By McCann's Theorem (\cite{deb2023multivariate,mccann1995existence,villani2021topics}), $R = (F_1, \ldots, F_d)^\T$, $\tilde R = (\tilde F_1, \ldots, \tilde F_d)^\T$ and $W = (F_1^{-1} \circ \tilde F_1, \ldots, F_d^{-1} \circ \tilde F_d)$.
Then, it is easy to see that Assumptions (A) and (C) hold.
Assumption (B) holds if each $F_j$ and $\tilde F_j$ admit finite moment generating functions of order $\alpha > 0$.
As for Assumption (D), it holds automatically since both $\mathbb J_W$ and $\mathbb J_R$ are diagonal.
Finally, one can show that $f(\tau) \leq L^2 \tau^2 \E \|\mb \epsilon\|^2$ which goes to 0 if we choose $\tau = \tau_n = o(n^{-1/2})$.
Note that this example motivates many choices of $F_{\U}$ and $F_{\mb \epsilon}$.
For example, we can choose $F_{\U} =\mc U^d$ and $F_{\mb \epsilon} = N(\mb 0_d, \mb I_d)$.

\end{example}

\section{Technical Lemmas for Proving Theorem \ref{theorem: rank preservation}}\label{appendix: proof of rank preservation}

\begin{lemma}[Rank Preservation]\label{lemma: rank preservation of optimal transport}
Under Assumptions (A)-(D), the optimal transport map $W$ exactly preserves the population ranks of $(\tilde \U_i)_{i=1}^n$, almost surely under $F_{\tilde \U}$.
Furthermore, $(\V_i)_{i = 1}^n$ approximately preserves the empirical ranks of $(\tilde \U_i)_{i = 1}^n$.
That is, $\|\mb R^{\V} - \tilde{\mb R}\|_n^2 = O_p\paren{r_{n, d}^* \paren{\log n}^{t_{\alpha, d}}}$, where $r_{n, 1}^* = 0$, and $r_{n, d}^* = r_{n, d}$ when $d > 1$.
\end{lemma}

\noindent{\bf Proof of Lemma \ref{lemma: rank preservation of optimal transport}:} 
To demonstrate the preservation of population ranks, note that $R$ and $W$ are the optimal transportation maps. By the Brenier-McCann Theorem, see Theorem 2.2 of \cite{ghosal2022multivariate}, $R = \nabla \Phi_1$ and $W = \nabla \Phi_2$ for some real-valued convex functions $\Phi_1$ and $\Phi_2$ on $\mc R^d$ under Assumptions (A).
Thus, $\mathbb J_R$ and $\mathbb J_W$ are positive semidefinite.
By the chain rule and Assumption (D),
$$
J_{R \circ W}\paren{\tilde \U} = J_{R \circ W}\paren{\tilde \U}^\T,
$$
almost surely, which implies that $R \circ W$ is the gradient of some real-valued function $\Phi$, i.e., $R \circ W = \nabla \Phi$.
Moreover, since both $\mathbb J_R$ and $\mathbb J_W$ are symmetric and positive semidefinite, we know that $\mathbb J_{R \circ W}\paren{\tilde \U}$, or the Hessian of $\Phi$, is also positive semidefinite.
To this end, we conclude that $\Phi$ is a convex function, and by the Brenier-McCann Theorem, $R \circ W(\tilde \U) = \tilde R(\tilde \U)$, almost surely, meaning that population ranks of $(\mb V_i)_{i = 1}^n$ are the same as those of $(\tilde \U_i)_{i = 1}^n$.

Next, we work on the preservation of empirical ranks by bounding $\|\mb R^{\V} - \tilde{\mb R}\|_n^2$.
When $d = 1$, the definition of multivariate ranks coincides with the univariate ranks by a normalizing constant.
By definition, the ranks of $(\V_i)_{i = 1}^n$ are the same as those of $(\tilde \U_i)_{i = 1}^n$ since $W = F_{\U}^{-1} \circ F_{\tilde \U}$, the optimal transport that pushes $F_{\tilde \U}$ to $F_{\U}$, is increasing, resulting $\|\mb R^{\V} - \tilde{\mb R}\|_n^2 = 0$.
For the rest of the proof, we focus on the case of $d \geq 2$. Note that $\|\bm R^{\V} - \tilde{\bm R}\|^2_n $ is bounded by 
\begin{align*}
& C_1 n^{-1}\sum_{i = 1}^n \|\bm R_i^{\V} - R(\V_i)\|^2 + C_2 n^{-1}\sum_{i = 1}^n \|R(\V_i) - \tilde R(\tilde \U_i)\|^2 \\
& + C_3 n^{-1}\sum_{i = 1}^n \|\tilde{\mb R}_i - \tilde R(\tilde \U_i)\|^2 \equiv \text{I + II + III},
\end{align*}
where $C_1$, $C_2$, and $C_3$ are universal constants. Under Assumptions (A)-(C), by Theorem 2.2 of \cite{deb2021rates},
both I and III are $O_p\paren{r_{n, d} \times \paren{\log n}^{t_{\alpha, d}}}$.
From the first part of the proof, we have $\text{II} = 0$ almost surely.
Then, the desired result follows by combining I, II, and III. This completes the proof. \hfill $\square$

\begin{lemma}[Stability] \label{lemma: stability of empirical rank maps}
Under Assumptions (A)-(C), 
{\small 
\begin{eqnarray*}
\|\mb R^{\mb T} - \tilde{\mb R}\|_n^2 = O_p\big(\max(r_{n, d} \big(\log n^{t_{\alpha, d}}), \tau^2 n^{-1/2} + f(\tau)) \big).
\end{eqnarray*}
}
for $d \geq 1$.
\end{lemma}

\noindent{\bf Proof of Lemma \ref{lemma: stability of empirical rank maps}:} Note that 
$\|\bm R^{\mb T} - \tilde{\bm R}\|^2_n$ is bounded by 
\begin{align*}
& C_1 n^{-1}\sum_{i = 1}^n \|\bm R_i^{\mb T} - R(\U_{r(i)})\|^2 + C_2 n^{-1}\sum_{i = 1}^n \|\tilde{\mb R}_i - \tilde R(\tilde \U_i)\|^2\\
& + C_3 n^{-1}\sum_{i = 1}^n \|R(\U_{r(i)}) - \tilde R (\tilde \U_i)\|^2 \doteq \text{I + II + III}
\end{align*}
where $C_1$, $C_2$, and $C_3$ are all universal constants.
Since $\mb R^{\mb T}$ are also empirical ranks of $(\U_{r(i)})_{i = 1}^n$, we have $\text{I} = O_p\paren{r_{n, d} \paren{\log n}^{t_{\alpha, d}}}$ due to the consistency\footnote{The case when $d = 1$ is not proved in \cite{deb2021rates} due to simplicity, but it can be easily proved by using Komlós–Major–Tusnády approximation \cite{komlos1975approximation} to bound $\|\mb R^{\V} - \mb R^{\mb T}\|_n^2$ directly, resulting in $O_p(n^{-1})$.} of empirical ranks maps under Assumptions (A)-(C) (Theorem 2.2 of \cite{deb2021rates}).
Similarly, $\text{II} = O_p\paren{r_{n, d} \paren{\log n}^{t_{\alpha, d}}}$ as well.
As for III, note that (up to some constant factors),
\begin{align*}
\text{III} \quad & \leq n^{-1} \sum_{i = 1}^n \|R (\U_{r(i)}) - R (\tilde \U_i)\|^2 + \\
&  n^{-1} \sum_{i = 1}^n \|R(\tilde \U_i) - \tilde R (\tilde \U_i\|^2\\
& \leq L^2 \tau^2 n^{-1} \sum_{i = 1}^n \|\bm \varepsilon_i\|^2 + \sup_{\bm z \in \mc Z} \|R(\bm z) - \wt{R}(\bm z)\|^2\\
& = L^2 \tau^2 O_p(n^{-1/2}) + O(f(\tau)).
\end{align*}
As a result, III $= O_p(\tau^2 n^{-1/2} + f(\tau))$ with a rate depending on both $n$ and $\tau$.
Putting all terms together, 
we obtain the rate stated in Lemma \ref{lemma: stability of empirical rank maps}. Moreover,  if $f(\tau)$ goes to 0 as $\tau \to 0^+$ and let $\tau = \tau_n$ to control the rate of convergence for III, which, combined with the rates of I and II, 
can yield the rate of convergence of $\|\bm R^{\mb T} - \tilde{\bm R}\|^2_n$, as in Lemma \ref{lemma: rank preservation of optimal transport}.  This completes the proof. \hfill $\square$

\section{Technical Proofs of Main Results.}\label{appendix: proof of main results}
\noindent{\bf Proof of Lemma \ref{lem1}:} Let $\mb T \equiv (\mb T(\Z_i))_{i = 1}^n$;
$\pi_1 \equiv \pi_1(\mb T)$ and $\pi_2 \equiv \pi_2(\mb U)$ be permutation maps 
in that $\mb T_{\pi_1(i)}$ and $\U_{\pi_2(i)}$ have the rank $\mb h^d_i$;
$i = 1, \ldots, n$, where $(\mb h^d_i)_{i = 1}^n$ denotes the $d$-dimensional multivariate ranks 
defined by the Halton sequence, c.f., Appendix \ref{appendix: multivariate rank}.
Then, $r = \pi_2 \circ \pi_1^{-1}$ is the permutation map matching the ranks of $(\bm U_{r(i)})_{i = 1}^n$ and $(\bm Z_i)_{i = 1}^n$, where $\mb r = (r_1, \ldots, r_n)  \in \mb \Pi_n$, 
the set of all permutations of $(1, \ldots, n)$.
For any $\x, \mb z \in \mc R^d$, $\mb z \leq \x$ means $z_j \leq x_j$; $j = 1, \ldots d$.
Moreover, for convenience, denote $\Pr\big(\prod_{i = 1}^m\{\X_i \leq \x_i\}\big)$ as $\Pr\big(\mb \X_1 \leq \x_1, \ldots, \X_m \leq \x_m\big)$ for random vectors $(\X_i)_{i = 1}^m$ (not necessarily independent nor identically distributed), and fixed points $(\x_i)_{i = 1}^m \subset \mc R^d$.

\noindent{\textbf{Part (1):}} By \eqref{DPG}, we show that $(\bm U_{r(i)})_{i = 1}^n$ is 
i.i.d. following $F_{\U}$. Note that $(\pi_j(1),\ldots, \pi_j(n))$
and $(r(1), \ldots, r(n))$ are uniformly distributed over $\mb \Pi_n$; $j = 1,2$. For any 
$(\bm u_i)_{i = 1}^n \subset \mc R^d$, 
$\Pr\big(\prod_{j=1}^n \{\U_{r(j)} \leq \mb u_j\}\big)$ is $(n!)^{-1} \sum_{\mb r \in \mb \Pi_n} 
\Pr\big(\prod_{j=1}^n \{\U_{r(j)} \leq \mb u_j\}| \pi_1 = \mb r\big)$ since $\Pr\big(\pi_1 = \mb r\big)=(n!)^{-1}$.
Note that $r(j)=\pi_2 \circ \pi_1^{-1}(j)$. By Proposition 2.1 of \cite{deb2023multivariate}, $(\mb U_{r(j)})_{j = 1}^n$ and $\pi_2$ are
independent\footnote{A multivariate version of the fact that order statistics and rank statistics are independent.}, given $\pi_1 = \mb r$,
$\Pr\big(\prod_{j=1}^n \{\U_{\pi_2 \circ \pi_1^{-1}(j)} \leq \mb u_j\} | \pi_1 = \mb r\big)$
equals to $\Pr\big(\prod_{j=1}^n \{\U_{\pi_2 \circ \pi_1^{-1}(j)} \leq \mb u_j\} | \pi_1 = \mb r, \pi_2 = \mb r\big)$,
which is $\Pr\big(\prod_{j=1}^n \{\U_{j} \leq \mb u_j\} | \pi_1 = \mb r, \pi_2 = \mb r\big)$. 
By the independence of $(\U_j)_{j = 1}^n$ and $\pi_1$ given $\pi_2 = \mb r$, 
$\Pr\big(\prod_{j=1}^n \{\U_j \leq \mb u_j\} | \pi_1 = \mb r, \pi_2 = \mb r\big)$ is
$ \Pr\big(\prod_{j=1}^n \{\U_j \leq \mb u_j\} | \pi_2 = \mb r\big)$. 
Hence, $\Pr\big(\prod_{j=1}^n \{\U_{r(j)} \leq \mb u_j\}\big)$ is equal to 
$(n!)^{-1} \sum_{\mb r \in \mb \Pi_n}
\Pr\big(\prod_{j=1}^n \{\U_j \leq \mb u_j\}| \pi_2 = \mb r\big)$, 
which is $\Pr\big(\prod_{j=1}^n \{\U_j \leq \mb u_j\}\big)$ since $\Pr\big(\pi_2 = \mb r\big) = (n!)^{-1}$ for any $\mb r \in \mb \Pi_n$.
This implies that $(\U_{r(i)})_{i = 1}^n$ is 
a random sample from $F_{\U}$, which in turn implies the desired result.

\noindent{\textbf{Part (2):}} Note that $H(\Z^\prime) = H(\Z^\prime_1, \ldots, \Z^\prime_n)$, where $\Z^{\prime}_i=\tilde G(W(\U_{r(i)};\e_i))$, $\mb U = (\mb U_i)_{i = 1}^n$, and $\mb e = (\mb e_i)_{i = 1}^n$ are random samples from $F_{\U}$ and $F_e$. Let $\X \deq \Y$ denote $\X$ and $\Y$ having the same distribution.
Furthermore, $\mb Z^\prime$ is only related to $\Z$ through $\pi_1 = \pi_1(\mb T(\Z))$. Then,
$H(\mb Z^\prime) | \Z$ has the same distribution of $H\big(\tilde G(W(\U_{r(1)};\e_1)), \ldots, \tilde G(W(\U_{r(n)};\e_n))\big) | \pi_1$, which is $H\big(\vphantom{U_{r(1)}} \tilde G(W(\U_1;\e_1)), \ldots, \tilde G(W(\U_n;\e_n))\big) | \pi_1, r = (1, \ldots, n)$ by the conditional independence of $(r(i))_{i = 1}^n$ and $(\mb U_{r(i)})_{i = 1}^n$ 
given $\mb Z$, c.f., Proposition 2.1 of \cite{deb2023multivariate}. By the permutation invariance property of 
$H(\cdot)$, it reduces to $H\big(\vphantom{U_{r(1)}} \tilde G(W(\tilde{\U}_1;\tilde{\e}_1)), \ldots, \tilde G(W(\tilde{\U}_n;\tilde{\e}_n))\big) \deq H(\mb Z^\prime)$ by the result of \textbf{Part (1)} that $\mb Z^\prime$ is i.i.d. from 
$\tilde F_{\Z}$, which is constructed using independent random samples $(\tilde{\mb U}_i)_{i = 1}^n$
from $F_{\U}$ and $(\tilde{\mb e}_i)_{i = 1}^n$ from $F_{\e}$. 
Therefore, $H(\mb Z^\prime)$ and $\mb Z$ are independent.
This concludes the proof. \hfill $\square$

\noindent{\bf Proof of Theorem \ref{thm1}:} From \eqref{DPG}, $\Z^{\prime(d)}$ is a conditionally independent sample 
of size $n$ given $\Z$ following $F_{\Z}$.
Let $X=I(H(\Z^{\prime}) \leq x)$ and $X^{(d)}=I(H(\Z^{\prime(d)}) \leq x) \in [0, 1]$ for any $x \in \mc R$.
By Hoeffding's Lemma, $\E_{\Z^{\prime(d)} \mid  \Z} \exp(s(X^{(d)} - \E_{\Z^{\prime(d)} 
\mid \Z} X^{(d)})) \leq \exp(s^2/8)$ a.s. for any $s>0$, where $\E_{\Z^{\prime} \mid \Z}$ is the conditional expectation with respect to $\Z^{\prime(d)}$ given $\Z$.
By Markov's inequality and the conditional independence between $\Z^{'(1)}, \ldots, \Z^{\prime(d)}$ given $\Z$, for any $t>0$ and $s=4 t$,
{\small 
\begin{align*}
& \Pr\Big(\abs{\tilde F_{H^{'}}(\bm x) - \tilde F_H(\bm x)} \geq t\Big) \\
& = \Pr\Big(\big|D^{-1} \sum_{d = 1}^D X^{(d)} - \E_{\Z^{\prime(d)} \mid \Z} X^{(d)}\big| \geq t\Big)\\
 & \leq \exp(-s D t) 
\E_{\Z} \prod_{d=1}^D \E_{\Z^{\prime(d)} \mid \Z} \big(
\exp\big(s(X^{(d)} - \E_{\Z^{\prime(d)} \mid \Z} X^{(d)})\big) \big)
\\
 & \leq \exp(-s D t) \exp(D s^2 / 8) \leq \exp(-2 D t^2).
\end{align*}
}
For any $\delta \in (0, 1)$, by choosing $t = \sqrt{\frac{\log \frac{4}{\delta}}{2D}}$, we have $\abs{\tilde F_{H^{'}}(\bm x) - \tilde F_H(\bm x)} \leq \sqrt{\frac{\log \frac{4}{\delta}}{2D}}$, with probability at least $1 - \frac{\delta}{2}$. On the other hand, $\abs{\tilde F_H(\bm x) - F_H(\bm x)} \leq 
\text{KS}(\tilde F_H,F_H) \leq \text{TV}(\tilde{\Z}, , \Z)$ with probability at least $1 - \frac{\delta}{2}$, where $\text{TV}(\tilde{\Z}, , \Z)$ is the total variation distance between the distributions 
of $\tilde{\Z}$ and $\Z$. Using the union bound to combine these foregoing results, we obtain that  
\begin{equation*}
    \sup_{\bm x} \abs{\hF_H(\bm x) - F_H(\bm x)} \leq \sqrt{\frac{\log \frac{4}{\delta}}{2D}} + 
\text{TV}(\tilde{\Z}, \Z),
\end{equation*}
with probability at least $1 - \delta$. Note that $\text{TV}(\tilde{\Z}, \Z)=
|\mathbb S| \cdot \text{TV}(\hFz, \Fz)$. This completes the proof. \hfill $\square$

\noindent{\bf Proof of Theorem \ref{thm2}:} In the setting of Theorem \ref{thm1}, note that $H(\Z)=T( \hat{\bm \theta},\bm \theta)$.
By Theorem 3 of \cite{shen2022data},  the conditional distribution of 
$H^{'}=H(\Z^{\prime})=T(\hat{\bm \theta}^{'(d)}, \hat{\bm \theta})$ given $\Z$ remains the 
same as the distribution of $T(\bm \theta, \hat{\bm \theta})$ for any $\Z$, that is,
for any event $\bm x \in \mc R^d$, 
\begin{eqnarray*}
& F_{H|\Z}(\bm x)=\Pr(T(\hat{\bm \theta}^{'(d)}, \hat{\bm \theta})\leq \bm x \mid \Z) \\
& =\Pr(T(\hat{\bm \theta}, \bm \theta) \leq \bm x)=F_{H}(\bm x); d=1,\ldots,D, 
\end{eqnarray*}
implying that $Err_{H}=0$. This completes the proof. 
\hfill $\square$

\section{Learning Theory for Normalizing Flows}\label{appendix: learning theory}

This section will leverage the maximum likelihood theory to derive novel results for normalizing flows. Initially, we present Lemma \ref{lem2}, which a variant of Theorem 1 of
\cite{dai2022coupled} that is a penalized  version of Theorem 1 of \cite{wong1995probability}.

Let $\mathcal{F}_j=\{G \in \mathcal F: P(G) \leq j\}$. Define the metric entropy of $\mathcal{F}_k$ under the Hellinger distance $h(\cdot,\cdot)$. The approximation error for $\pz$ is $\gamma_{n_h}=\rho^{1/2}_{\alpha}(\pz, \hpz) =  
\Big(\mathbb{E} g_\alpha \big(\frac{\pz}{\hpz}\big)\Big)^{1/2}$, where
$g_{\alpha}(x)=\alpha^{-1}(x^{\alpha}-1)$ for $\alpha \in (0,1)$, which a slightly stronger metric than the Kullback-Leibler divergence; c.f., 
Section 4 of \cite{wong1995probability}.

  Lemma \ref{lem2} establishes the estimation
error for the distribution under the total variation norm $\text{TV}(\hFz, \Fz)$ given the fact that the total variation distance is upper bounded by the corresponding Hellinger distance
in that $\text{TV}(\hFz, \Fz) \leq h(\hFz,\Fz)$.

\begin{lemma} (Estimation error of $F_{\Z}=F_{\V} \circ \tilde{G}^{-1}$ by PASS in \eqref{mle})
\label{lem2}
Assume that $\tilde{\bm Z}=(\tilde{\bm Z}_i)_{i=1}^{n_h}$ is random sample following $F_{\Z}$
and $\tilde{G}$ is the regularized MLE obtained from \eqref{mle}.
Suppose there exist some positive constants $C_1$-$C_2$, such that, for any $\epsilon_n >0$, and $\lambda \geq 0$,
{\small 
\begin{align}
\label{eqn:entropy_direct}
& \sup_{j \geq 1} \int_{2^{-8}L_j}^{2^{1/2}L_j^{1/2}} H^{1/2} \big(u/C_2, {\cal F}_j\big) du/ L_j \leq C_3 n_h^{1/2}, 
\end{align}
}
where $L_j = C_1 \epsilon_n^2 + \lambda(j-1)$. Then
$F_{\Z}=F_{\bm V}(\tilde G^{-1})$ satisfies:
{\small 
\begin{equation*}
P \big(h(F_{\Z}, \Fz)
\geq \eta_n \big) \leq 8 \exp(-C_4 n_h \eta_{n_h}^2), 
\end{equation*}                    
}
with $\eta_{n_h} = \max \big(\epsilon_{n_h}, \gamma_{n_h} \big)$
provided that $\lambda$ is tuned: $\lambda \max (P(G^*), P(G),1) \leq C_5 \eta_{n_h}^2$, $G^*
\in \mathcal F $ is an approximating point of $G$, and $C_4>0$ is a constant.
Hence, $h(\hFz, \Fz)=O_p(\eta_{n_h})$ as $n \rightarrow \infty$ under $\pz$.
\end{lemma}

Now consider RealNVP flow \cite{dinh2016density}, a normalizing flow with $K$ affine-coupling layers with the reverse order permutation we used in Section \ref{nlp}. Let $\z_k=[\z_k^{(1)}, \z_k^{(2)}]$ be the output at $k$-th layer from its input
$\z_{k - 1}=[\z_{k-1}^{(1)}, \z_{k-1}^{(2)}]$, and $[\bm t_k, \bm s_k] = f_{\bm \theta_k}(\z_{k-1}^{(1)})$ is a fully connected neural network parametrized by weights $\bm \theta_k$;  $k=1,\cdots$. Then, the coupling flow yields an output $T_{\bm \theta}(\z_0)=T_{\bm \theta_K} \circ \ldots \circ T_{\bm \theta_1}$ from an input
$\z_0=[\z_0^{(1)}, \z_0^{(2)}]$ (can be thought of as $\mb V_i$'s, the latent representation in \eqref{DPG}) through function composition: $\bm z_k  =  T_{\bm \theta_{K}}(\z_{k-1})$, defined recursively, via
{\small
\begin{eqnarray}
\label{k-flow}
\bm z_k=\brac{\z_{k-1}^{(1)}, \z_{k-1}^{(2)} \odot \exp(\bm s_k) + \bm t_k} P; k=1,\cdots,K,
\end{eqnarray}}
where $P$ is a $2d \times 2d$ permutation matrix whose entries are either 0 or 1 with the rows or columns arranged in the reverse order of the identity matrix, and $\odot$ denotes the componentwise product. Note that 
each $f_{\bm \theta_k}$ approximates a scale function $\bm s_k$ and a translation function
$\bm t_k$ for $\z_{k-1}^{(2)}$, while $T_{\bm \theta_{k}}$ approximates a smooth and invertible
function. Assume that $d$ is independent of the sample size $n_h$.
Let $\mb \theta \equiv (\bm \theta_1, \dots, \mb \theta_K) \in \mb \Theta \equiv \mb \Theta_1 \times \dots \times \mb \Theta_K$ represent the parameter of the normalizing flow. 
Denote $\mc T_{\mb \Theta}$ as the class of $K$-layer affine coupling flow defined above parametrized by $\mb \Theta$.
Using the framework of \eqref{mle}, we obtain that

{\footnotesize
\begin{align}
\tilde G = \argmax_{G_{\mb \theta} \in \mc T_{\mb \Theta}} &
n_h^{-1} \sum_{i = 1}^{n_h} \log \paren{\pv(G_{\mb \theta}^{-1}(\tilde{\mb Z}_i)) \cdot |\det J_{G_{\mb \theta}}(G_{\mb \theta}^{-1}(\tilde{\mb Z}_i))|^{-1}} \nonumber\\
& \hphantom{used for identation} + \lambda P(G_{\mb \theta}) \label{mle1}
\end{align}}
based on a holdout sample $(\tilde{\mb Z}_i)_{i = 1}^{n_h}$ from $\Fz$, the base density $\pv$,
and the penalty function $P(G_{\mb \theta})=\|\bm \theta\|^2_2$.

The following conditions are assumed:

\begin{assumption}[Transport Approximation]\label{assumptionF1}
The transport $G = T_{\bm \theta}$ can be
approximated by a normalizing flow with an approximation error
$\gamma_{n_h}=\rho^{1/2}_{\alpha}(\pz, \hpz)$ defined by the distance
between the true density $\pz$ of $\Z$ and
the best approximating density $\hpz$ induced by the flow, for some $0<\alpha<1$. 
\end{assumption}

\begin{assumption}[Neural Networks]\label{assumptionF2}
The fully connected neural networks retain the same structure for each layer of the flow, with a parameter space
{\small
\begin{equation*}
      \bm{\Theta}_k= \left\{\bm \theta_k:
    \max(\|\bm W_{k, l}\|_{\infty}, \|\bm b_{k, l}\|_{\infty}) \leq 1,
\ \|\bm \theta_k\|_0 \leq s_k \right\},
\end{equation*}}
where $\bm \theta_k=(\bm W_{k, l},\bm b_{k, l})_{1 \leq l \leq h}$ are the network weight matrix and bias vector at flow layer $k$ and neural network layer $l$,
$\|\bm \theta_k\|_0=\sum_{l=0}^L (\|\bm W_{k, l}\|_0 + \|\bm b_{k, l}\|_0)$ is the $L_0$-norm and $\|\cdot\|_{\infty}$ denotes
the $L_{\infty}$-norm. For this network, the first layer consists of $d$ nodes and equal width for the middle and the layer consists of $2 d$ layers. The total number of nonzero network parameters in the $K$-layer affine coupling flow is $s = \sum_{k = 1}^K s_k$.    
\end{assumption}

\begin{assumption}[Base Density]\label{assumptionF3}
The density of $\V$ $\pv(\mb v)$ satisfies
the Lipschitz condition with a Lipschitz constant $L_p$ with respect to $\ell_2$ norm.
and $B_p$ is the sup-norm of $\pv(\mb v)$.    
\end{assumption}

\begin{lemma}
\label{lam:4}
Under Assumptions F1-F3, for any $\mb \theta, \mb \theta^\prime \in \mb \Theta$, $\|p_{\mb \theta} - p_{\mb \theta^\prime}\| \leq \sum_{k = 1}^K A_k \|f_{\mb \theta_k} - f_{\mb \theta_k^\prime}\|$, where
{\small
$$
A_k = \exp(B_f d K) \big(L_p \paren{C(B_f) \sqrt{d}}^k \prod_{m = 1}^{k - 1} (\tau_m \vee 1) + B_p C(B_f) d \tau_k \big),
$$}
where $B_f$ is the $\ell_\infty$ bound on both input vectors and output vectors of $T_{\bm \theta_k}$ and $f_{\mb \theta_k}$ for all $k = 1, \ldots, K$; $C(B_f)$ is a universal constant depending on $B_f$ only; $\tau_k$ is the Lipschitz constant of $f_{\mb \theta_k}$ with respect to $\ell_2$ norm in both input and output space.
\end{lemma}

\noindent\textbf{Proof of Lemma \ref{lam:4}:}
Let $T = T_K \circ \dots \circ T_1$ where $T_k \equiv T_{\mb \theta_k}$; $k = 1, \dots, K$. Denote $\mb z_k = [\mb z_k^{(1)}, \mb z_k^{(2)}]$ as the corresponding output of $k$-th flow layer, and let $\mb v = T^{-1}(\mb z)$ be the input of the flow, and $\mb z$ is the final output; Let $[\mb s_k, \mb t_k] = f_{\mb \theta_k}\paren{\mb z_{k - 1}^{(1)}}$ be the scaling and translation parameters of $k$-th flow layer defined by a neural network $f_{\mb \theta_k}$; Define $G_k = T_k \circ \dots \circ T_1$; $k = 1, \dots, K$.
Define $T^\prime$, $T_k^\prime$, $\mb z_k^\prime$, $\mb v^\prime$, $[\mb s_k^\prime, \mb t_k^\prime]$ and $G_k^\prime$ similarly, but with another set of parameters $\mb \theta^\prime \in \mb \Theta$. 
Let $p(\mb v) \equiv p_{\small \mb V}(\mb v)$ be the base density and $p_{\mb \theta}(\mb z)$ be the parametrized density of the output of $\mb z$. 
For this proof, we will use $C(B_f)$ to indicate a function that only depends on constant $B_f$, which is not necessarily the same every time it appears, but remains a universal constant of $B_f$ in any case.

We begin with the following decomposition,
{\small 
\begin{align*}
&\quad \abs{p_{\mb \theta}(\mb z) - p_{\mb \theta^\prime}(\mb z)}\\
= & \quad \abs{p(\mb v) \cdot \abs{\det J_T(\mb v)}^{-1} - p(\mb v^\prime) \cdot \abs{\det J_{T^\prime}(\mb v^\prime)}^{-1}}\\
\leq & \quad \abs{p(\mb v) - p(\mb v^\prime)} \cdot \abs{\det J_T(\mb v)}^{-1} \\
& + p(\mb v^\prime) \cdot \abs{\abs{\det J_T(\mb v)}^{-1} - \abs{\det J_{T^\prime}(\mb v^\prime)}^{-1}}\\
\leq & \quad L_p \|\mb v - \mb v^\prime\|_2 \cdot \abs{\det J_T(\mb v)}^{-1} \\
& + B_p \cdot \abs{\abs{\det J_T(\mb v)}^{-1} - \abs{\det J_{T^\prime}(\mb v^\prime)}^{-1}}\\
\leq &\quad  L_p \exp\paren{KB_f d} \|\mb v - \mb v^\prime\|_2\\
& + B_p \cdot \abs{\abs{\det J_T(\mb v)}^{-1} - \abs{\det J_{T^\prime}(\mb v^\prime)}^{-1}}.
\end{align*}
}

Next, we bound $A=\|\mb v - \mb v^\prime\|_2$ and $B =\abs{\abs{\det J_T(\mb v)}^{-1} - \abs{\det J_{T^\prime}(\mb v^\prime)}^{-1}}$ separately.

For $A$, note that
{\small 
\begin{align*}
A = & \quad\|\mb v - \mb v^\prime\|_2 = \|T_1^{-1}(\mb z_1) - (T_1^\prime)^{-1}(\mb z_1^\prime)\|_2\\
\leq & \quad\|T_1^{-1}(\mb z_1) - T_1^{-1}(\mb z_1^\prime)\|_2 + \|T_1^{-1}(\mb z_1^\prime) - (T_1^\prime)^{-1}(\mb z_1^\prime)\|_2\\
\leq & \quad C(B_f) \sqrt{d} (\tau_1 \vee 1) \cdot \|\mb z_1 - \mb z_1^\prime\|_2\\
& + C(B_f) \sqrt{d} \cdot \|f_{\mb \theta_1}(\mb z_1^{\prime(1)}) - f_{\mb \theta_1^\prime}(\mb z_1^{\prime(1)})\|_2,
\end{align*}
}
where we use the Lipschitz property of $T_1^{-1}$ for the first term, and the difference between $T_1^{-1}$ and $(T_1^\prime)^{-1}$ for the second term. See Lemmas \ref{lemmaA1} and \ref{lemmaA2} for the derivation of these two bounds.
Then a deduction can be done to bound $ \|\mb z_1 - \mb z_1^\prime\|_2$ in a similar way as above, which gives the final bound as follows,
\begin{align*}
A & \leq \sum_{k = 1}^K \paren{C(B_f) \sqrt{d}}^k \prod_{m = 1}^{k - 1} \paren{\tau_m \vee 1} \|f_{\mb \theta_k}(\mb z_k^{\prime(1)}) - f_{\mb \theta_k^\prime}(\mb z_k^{\prime(1)})\|_2\\
& \leq \sum_{k = 1}^K \paren{C(B_f) \sqrt{d}}^k \prod_{m = 1}^{k - 1} \paren{\tau_m \vee 1} \|f_{\mb \theta_k} - f_{\mb \theta_k^\prime}\|_\infty.
\end{align*}

For $B$,  note that 
{\small 
\begin{align*}
\abs{\det J_T(\mb v)}^{-1} = & \exp(- \mb 1^\T \mb s_K) \cdot \abs{\det J_{G_{K - 1}}(\mb v)}^{-1}\\
= & \cdots = \exp(- \mb 1^\T(\mb s_K + \cdots + \mb s_1)).
\end{align*}
}
Then,
{\small 
\begin{align*}
B = & \quad\big|\abs{\det J_T(\mb v)}^{-1} - \abs{\det J_{T^\prime}(\mb v^\prime)}^{-1} \big|\\
\leq & \quad\exp(- \mb 1^\T \mb s_K) \cdot \Big|\abs{\det J_{G_{K - 1}}(\mb v)}^{-1} - \abs{\det J_{G_{K - 1}^\prime}(\mb v^\prime)}^{-1}\Big|\\
& + \abs{\det J_{G_{K - 1}^\prime}(\mb v^\prime)}^{-1} \cdot \abs{\exp(- \mb 1^\T \mb s_K) - \exp(- \mb 1^\T \mb s_K^\prime)}\\
\leq & \quad\exp(dB_f) \cdot \abs{\abs{\det J_{G_{K - 1}}(\mb v)}^{-1} - \abs{\det J_{G_{K - 1}^\prime}(\mb v^\prime)}^{-1}}\\
& + \exp(d(K-1)B_f) \cdot \abs{\exp(- \mb 1^\T \mb s_K) - \exp(- \mb 1^\T \mb s_K^\prime)}.
\end{align*}
}
To bound the second term on RHS, we first notice that $\exp(- \mb 1^\T \mb s)$ is $\sqrt{d} \exp(dB_f)$-Lipschitz 
in the $\|\cdot\|_2$-norm, which gives us, for any $\mb s_K, \mb s_K^\prime \in \mc R^d$,
$$
\abs{\exp(- \mb 1^\T \mb s_K) - \exp(- \mb 1^\T \mb s_K^\prime)} \leq \sqrt{d} \exp(dB_f) \cdot \|\mb s_K - \mb s_K^\prime\|_2.
$$
It remains to bound $\|\mb s_K - \mb s_K^\prime\|_2$, which we proceed as follows, with a general $k = 1, \dots, K$ instead:
{\small 
\begin{align*}
& \quad \|\mb s_k - \mb s_k^\prime\|_2\\
= & \quad\|f_{\mb \theta_k}(\mb z_{k - 1}^{(1)}) - f_{\mb \theta_k^\prime}(\mb z_{k - 1}^{\prime(1)})\|_2\\
\leq & \quad\|f_{\mb \theta_k}(\mb z_{k - 1}^{(1)}) - f_{\mb \theta_k}(\mb z_{k - 1}^{\prime(1)})\|_2 + \|f_{\mb \theta_k}(\mb z_{k - 1}^{\prime(1)}) - f_{\mb \theta_k^\prime}(\mb z_{k - 1}^{\prime(1)})\|_2\\
\leq & \quad\tau_k \|\mb z_{k - 1}- \mb z_{k - 1}^{\prime}\|_2 + \|f_{\mb \theta_k}(\mb z_{k - 1}^{\prime(1)}) - f_{\mb \theta_k^\prime}(\mb z_{k - 1}^{\prime(1)})\|_2\\
\leq & \quad\tau_k C(B_f) \sqrt{d} \|f_{\mb \theta_k}(\mb z_k^{(1)}) - f_{\mb \theta_k^\prime}(\mb z_k^{(1)})\|_2 + \|f_{\mb \theta_k}(\mb z_{k - 1}^{\prime(1)}) - f_{\mb \theta_k^\prime}(\mb z_{k - 1}^{\prime(1)})\|_2,
\end{align*}
}
where we use Lemma \ref{lemmaA2} again for the last inequality. After cleaning and arranging some terms, we obtain the bound
\begin{align*}
B \leq & \quad \exp(dB_f) \cdot \abs{\abs{\det J_{G_{K - 1}}(\mb v)}^{-1} - \abs{\det J_{G_{K - 1}^\prime}(\mb v^\prime)}^{-1}}\\
& + d\exp(dKB_f)\tau_K C(B_f) \cdot \|f_{\mb \theta_K} - f_{\mb \theta_K^\prime}\|_\infty,
\end{align*}
which, by deduction on $\abs{\abs{\det J_{G_{K - 1}}(\mb v)}^{-1} - \abs{\det J_{G_{K - 1}^\prime}(\mb v^\prime)}^{-1}}$, gives the following bound
$$
B\leq C(B_f) d \exp(KdB_f) \cdot \sum_{k = 1}^K \tau_k \|f_{\mb \theta_k} - f_{\mb \theta_k^\prime}\|_\infty.
$$

Finally, we combine bounds on $A$ and $B$ to obtain the desired bound in Lemma \ref{lam:4}. \hfill $\square$

Next we present Lemma \ref{lemmaA1} and \ref{lemmaA2}  used in the proof of Lemma \ref{lam:4}, along with their proofs.
\begin{lemma}\label{lemmaA1}
    Consider one affine coupling layer $f: [\mb x^{(1)}, \mb x^{(2)}] \in \mc R^{2d} \mapsto [\mb x^{(1)}, \mb x^{(2)} \odot \exp(s) + t] P \in \mc R^{2d}$ where $[s, t] = g(\mb x^{(1)})$ is a neural network mapping from $\mc R^d$ to $\mc R^{2d}$, with Lipschitz constant $\tau > 0$ w.r.t. $\|\cdot\|_2$. Let $B_f > 0$ be the $\ell_\infty$ bound on both input and output vectors of $f$ and $g$. Then for any $\mb z, \mb y \in \mc R^{2d}$, we have $\|f^{-1}(\mb z) - f^{-1}(\mb y) \|_2 \leq C(B_f) \sqrt{d} (\tau \vee 1) \cdot \|\mb z - \mb y\|_2$.
\end{lemma}
\noindent\textbf{Proof of Lemma \ref{lemmaA1}:} 
By definition of $f$, 
{\small 
\begin{align*}
& \|f^{-1}(\mb z) - f^{-1}(\mb y)\|_2 \leq \quad \|\mb z^{(1)} - \mb y^{(1)}\|_2\\
& + \|\mb z^{(2)} \odot \exp(-s(\mb z^{(1)})) - \mb y^{(2)} \odot \exp(-s(\mb y^{(1)}))\|_2\\
& + \|t(\mb z^{(1)}) \odot \exp(-s(\mb z^{(1)})) - t(\mb y^{(1)}) \odot \exp(-s(\mb y^{(1)}))\|_2.
\end{align*}
}
The second term on RHS can be bounded:
{\small 
\begin{align*}
&\quad \|\mb z^{(2)} \odot \exp(-s(\mb z^{(1)})) - \mb y^{(2)} \odot \exp(-s(\mb y^{(1)}))\|_2 \\
\leq& \quad \|\mb z^{(2)}\|_2 \cdot \|\exp(-s(\mb z^{(1)})) - \exp(-s(\mb y^{(1)}))\|_2\\
& + \|\mb z^{(2)} - \mb y^{(2)}\|_2 \cdot \|\exp(-s(\mb y^{(1)}))\|_2\\
\leq & \quad \sqrt{d} B_f \exp(B_f) \cdot \|s(\mb z^{(1)}) - s(\mb y^{(1)})\|\\
& + \sqrt{d}\exp(B_f) \cdot \|\mb z^{(2)} - \mb y^{(2)}\|_2\\
\leq & \quad \sqrt{d} B_f \exp(B_f) \tau \cdot \|\mb z^{(1)} - \mb y^{(1)}\|_2\\
& + \sqrt{d} \exp(B_f) \cdot \|\mb z^{(2)} - \mb y^{(2)}\|_2.
\end{align*}
}
The third term on RHS can be bounded: 
\begin{align*}
& \quad \|t(\mb z^{(1)}) \odot \exp(-s(\mb z^{(1)})) - t(\mb y^{(1)}) \odot \exp(-s(\mb y^{(1)}))\|_2\\
\leq & \quad \|t(\mb z^{(1)})\|_2 \cdot \|\exp(-s(\mb z^{(1)})) - \exp(-s(\mb y^{(1)}))\|_2\\
& + \|\exp(-s(\mb y^{(1)}))\|_2 \cdot \|t(\mb z^{(1)}) - t(\mb y^{(1)})\|_2\\
\leq & \quad \sqrt{d} B_f \exp(B_f) \cdot \|s(\mb z^{(1)}) - s(\mb y^{(1)})\|_2\\
& + \sqrt{d} \exp(B_f) \cdot \|t(\mb z^{(1)}) - t(\mb y^{(1)})\|_2 \\
\leq & \quad \paren{\sqrt{d} B_f \exp(B_f) + \sqrt{d}\exp(B_f)} \cdot \|g(\mb z^{(1)}) - g(\mb y^{(1)})\|_2\\
\leq & \quad \paren{\sqrt{d} B_f \exp(B_f) + \sqrt{d}\exp(B_f)} \tau \cdot \|\mb z^{(1)} - \mb y^{(1)}\|_2.
\end{align*}
Combining all terms, we have the desired result in Lemma \ref{lemmaA1}, implying that the Lipschitz constant of $f^{-1}$ is $C(B_f)\sqrt{d}\paren{\tau \vee 1}$. \hfill $\square$

\begin{lemma}\label{lemmaA2}
    Consider two affine coupling layers $f_1$ and $f_2$ defined in the same way as in Lemma \ref{lemmaA1}, using neural networks $g_1$ and $g_2$ to generate scaling and translation parameters $[s_1, t_1]$ and $[s_2, t_2]$, respectively . Then for any $\mb y \in \mc R^{2d}$, we have $\|f_1^{-1}(\mb y) - f_2^{-1}(\mb y)\|_2 \leq C(B_f) \sqrt{d} \cdot \|g_1(\mb y^{(1)}) - g_2(\mb y^{(1)})\|_2$.
\end{lemma}
\noindent\textbf{Proof of Lemma \ref{lemmaA2}:}
Using the definition of $f_1$ and $f_2$, we begin with the following decomposition,
{\small
\begin{align*}
& \|f_1^{-1}(\mb y) - f_2^{-1}(\mb y)\|_2\\
\leq & \quad \|\mb y^{(2)} \odot \paren{\exp(-s_1(\mb y^{(1)})) - \exp(-s_2(\mb y^{(1)}))}\|_2\\
& + \|t_1(\mb y^{(1)}) \odot \exp(-s_1(\mb y^{(1)})) - t_2(\mb y^{(1)}) \odot \exp(-s_2(\mb y^{(1)}))\|_2.
\end{align*}}
Then, using a similar argument in the proof of Lemma \ref{lemmaA1}, the first term on RHS can be bounded by $\sqrt{d} B_f \exp(B_f) \cdot \|s_1(\mb y^{(1)}) - s_2(\mb y^{(1)})\|_2$, while the second term can be bounded by 
\begin{align*}
& \sqrt{d} B_f \exp(B_f) \cdot \|s_1(\mb y^{(1)}) - s_2(\mb y^{(1)})\|_2\\
& + \sqrt{d} \exp(B_f) \cdot \|t_1(\mb y^{(1)}) - t_2(\mb y^{(1)})\|,
\end{align*}
which, combined, give the desired bound. \hfill $\square$

In the literature, \cite{koehler2021representational} demonstrates how normalizing flows can be used as a universal approximator to complex distributions. As suggested by \cite{koehler2021representational}, even shallow affine coupling networks can approximate a smooth and invertible transport in Wasserstein distance well, as in GLOW. Yet, the approximation error bounds have not been available for essentially any flows.

\begin{prop} (Validity of PAI inference for the test \eqref{tests})
\label{prop1}
Under Assumptions \ref{assumptionF1}-\ref{assumptionF3}, the maximum likelihood transport
$\tilde G$ from \eqref{mle1} by RealNVP with the reverse permutation and a fully connected 
neural network satisfies:
\begin{eqnarray}
h(\hFz, \Fz) = O_p\paren{ \max\Big(\paren{\frac{s}{n_h} \log \frac{n_h}{s}}^{1/2}, \gamma_d\Big)},
\end{eqnarray}
where $\gamma_d$ is the approximation error and 
$s$ is the number of non-zero parameters in the neural network. 
In conclusion, by \eqref{main:thm:new}, the PAI test in \eqref{hypothesis1} is
valid for any inference sample $\mathbb S=(\bm X_i, Y_i)_{i=1}^{n_h}$
as the holdout size $n_h \rightarrow \infty$ and the MC size $D \rightarrow \infty$. 
\end{prop}

\noindent {\textbf{Proof of Proposition \ref{prop1}:} To apply Lemma \ref{lem2}, we bound the
metric entropy and compute the approximation error. 

\noindent \textit{Approximation Error.} By Assumption \ref{assumptionF3} and Lemma \ref{lam:4},  
$\gamma_{n_d} \leq C_5 \|p_{\theta^*}-p_{\theta^0}\|_{\infty} \leq C_5 \epsilon$ for some constant $C_5>0$, 
or approximation error $\gamma_d=\rho^{\alpha}_{0^+}(\pz, \hpz) =
C_5 \sqrt{d} \epsilon$, where $\rho^{\alpha}_{0^+}$ is the Kullback–Leibler divergence that
is upper bounded by the corresponding Hellinger distance when the densities are bounded away zero and
infinity.

\noindent \textit{Metric Entropy.}
Note that $\mathcal F_j=\{\bm \theta \in \mathcal F: \lambda P(G) \leq j, \|\bm \theta-\bm \theta^0\|_2 < 2 \epsilon\}$ is uniformly bounded.
Let $\mathcal{S}_{\infty}(u,m)$ be a $u$-cover of $\mathcal{F}_j$ in $\|\cdot\|_{\infty}$.
Define $g^+_k=g_k+u$ and $g^-_k=g_k-u$, where $g_k\in \mathcal{S}_{\infty}(u,m)$, $k=1,\ldots,m$.
Then $\{ g_1^{\pm},\ldots,g_m^\pm \}$ forms a $u$-bracket of $\mathcal{F}_j$.
Hence, $H(u,\mathcal{F}_j)=\min_m \log m \leq H_\infty(u,\mathcal{F}_j)$, where $H_\infty(u,\cdot)$ denotes the entropy under the sup-norm. Then,  by Lemma \ref{lam:4}, 
\begin{equation*}
    H_{\infty}(u,\mathcal{F}_j) \leq s \left(\log(\frac{\max(j,1)}{u})\right). 
\end{equation*}
Thus, the entropy integral in \eqref{eqn:entropy_direct} becomes
{\small 
\begin{equation*}
  \begin{split}
  & \int^{\sqrt{2}\epsilon}_{\epsilon^2/256} H^{1/2}_{B}(u/c_1,\mathcal{F}_j(A_j) ) du
  \leq \int^{\sqrt{2}\epsilon}_{\epsilon^2/256} \sqrt{s} \sqrt{\log(\frac{\max(j,1)}{u})}du  \\
  &\leq C_5 \sqrt{s} (\log(1/\epsilon))^{1/2}
\epsilon.
  \end{split}
\end{equation*}
}
We solve the integral equation:
$$
 C_5 \sqrt{s} (\log(1/\epsilon))^{1/2} \epsilon \leq C_6 \epsilon^2\sqrt{n_h},
$$
$\epsilon \asymp \sqrt{\frac{s}{n_h} \log \frac{n_h}{s}}$.
Hence, $\eta_{n_h}=\max(\epsilon,\gamma_d)$ by Lemma \ref{lem2}.
This completes the proof. \hfill $\square$

\bibliographystyle{ieeetr}
\bibliography{preref-1,diffpriv-1}


\end{document}